\documentclass{article}

\usepackage[final,nonatbib]{neurips_2018}

\usepackage[utf8]{inputenc} 
\usepackage[T1]{fontenc}    
\usepackage{hyperref}       
\usepackage{url}            
\usepackage{booktabs}       
\usepackage{amsfonts}       
\usepackage{nicefrac}       
\usepackage{microtype}      
\usepackage{lscape}
\usepackage{graphicx}
\usepackage{pgfplots}
\usepackage{todonotes}
\usepackage{multirow}

\usepackage{footmisc}

\usepackage{pifont}
\newcommand{\xmark}{\ding{55}}

\usepackage{amssymb}
\usepackage{amsmath}
\usepackage{dsfont}
\usepackage{amsthm}
\usepackage{color}

\usepackage{pgfplots}

\usepackage{mathtools}
\DeclarePairedDelimiterX{\infdivx}[2]{(}{)}{%
	#1\>\delimsize\|\>#2%
}
\newcommand{\KL}{KL\infdivx}
\newcommand{\kl}{kl\infdivx}
\newcommand{\klinv}{kl^{-1}}

\newcommand{\ii}{\mathds{1}}
\newcommand{\diag}{{\rm diag}}

\newtheorem{theorem}{Theorem}

\newcommand{\trainoutputs}{y_N}

\newtheorem{remark}[theorem]{Remark}

\title{Learning Gaussian Processes by\\Minimizing PAC-Bayesian Generalization Bounds}

\author{
 David Reeb\ \ \ \ \ \ \ \ \ \ \ \ Andreas Doerr\ \ \ \ \ \ \ \ \ \ \ \ Sebastian Gerwinn\ \ \ \ \ \ \ \ \ Barbara Rakitsch \\
 Bosch Center for Artificial Intelligence\thanks{\url{https://www.bosch-ai.com}}\\
Robert-Bosch-Campus 1\\
71272 Renningen, Germany \\
 \texttt{\{david.reeb,andreas.doerr3,sebastian.gerwinn,barbara.rakitsch\}@de.bosch.com}
}

\begin{document}

\maketitle

\renewcommand*{\thefootnote}{\arabic{footnote}}
\setcounter{footnote}{0}



\begin{abstract}
Gaussian Processes (GPs) are a generic modelling tool for supervised learning.
While they have been successfully applied on large datasets, their use in safety-critical applications is hindered by the lack of good performance guarantees. To this end, we propose a method to learn GPs and their sparse approximations by directly optimizing a PAC-Bayesian bound on their generalization performance, instead of maximizing the marginal likelihood. Besides its theoretical appeal, we find in our evaluation that our learning method is robust and yields significantly better generalization guarantees than other common GP approaches on several regression benchmark datasets. 

 %

\end{abstract}

\section{Introduction}\label{intro-section}

Gaussian Processes (GPs) are a powerful modelling method due to their non-parametric nature \cite{rasmussen-williams-book}. Although GPs are probabilistic models and hence come equipped with an intrinsic measure of uncertainty, this uncertainty does not allow conclusions about their performance on previously unseen test data. For instance, one often observes overfitting if a large number of hyperparameters is adjusted using marginal likelihood optimization \cite{bauer-understanding-sparse-GP-approximations}.
While a fully Bayesian approach, i.e.\ marginalizing out the hyperparameters, reduces this risk, it incurs a prohibitive runtime since the predictive distribution is no longer analytically tractable. Also, it does not entail out-of-the-box  safety guarantees.
 

In this work, we propose a novel training objective for GP models, which enables us to give \emph{rigorous} and quantitatively good performance guarantees on future predictions. Such rigorous guarantees are developed within Statistical Learning Theory (e.g.\ \cite{understanding-machine-learning-book}). But as the classical uniform learning bounds are meaningless for expressive models like deep neural nets \cite{understanding-deep-learning-requires-rethinking-generalization} (as e.g.\ the VC dimension exceeds the training size) and GPs or non-parametric methods in general, such guarantees cannot be employed for learning those models.
Instead, common optimization schemes are
(regularized) empirical risk minimization (ERM) \cite{understanding-deep-learning-requires-rethinking-generalization,understanding-machine-learning-book}, maximum likelihood (MLE) \cite{rasmussen-williams-book}, or variational inference (VI) \cite{jordan-ghahramani-et-al-vi,titsias09}. 

On the other hand, better non-uniform learning guarantees have been developed within the PAC-Bayesian framework \cite{mcallester-pac-bayesian-model-averaging,asdf2,catoni-thermodyn-of-ml} (Sect.\ \ref{framework-section}). They are specially adapted to probabilistic methods like GPs and can yield tight generalization bounds, as observed for GP classification \cite{seeger-classification-paper}, probabilistic SVMs \cite{tighter-pac-bayes-bounds-svm,pac-bayes-and-margins}, linear classifiers \cite{pac-bayes-learning-of-linear-classifiers}, or stochastic NNs \cite{nonvacuous-generalization-bounds-snn}. Most previous works used PAC-Bayesian bounds merely for the final evaluation of the generalization performance, whereas learning by optimizing a PAC-Bayesian bound has been barely explored \cite{pac-bayes-learning-of-linear-classifiers,nonvacuous-generalization-bounds-snn}. This work, for the first time, explores the use of PAC-Bayesian bounds {\it{(a)}} for GP training and {\it{(b)}} in the regression setting.


Specifically, we propose to learn full and sparse GP predictors $Q$ directly by minimizing a PAC-Bayesian upper bound $B(Q)$ from Eq.\ (\ref{pac-bayes-union-bound-eq}) on the true future risk $R(Q)$ of the predictor, as a principled method to ensure good generalization (Sect.\ \ref{learning-GPs-section}). Our general approach comes naturally for GPs because the KL divergence $KL(Q\|P)$ in the PAC-Bayes theorem can be evaluated analytically for GPs $P,Q$ sharing the same hyperparameters. As this applies to popular sparse GP variants such as DTC \cite{seeger-dtc}, FITC \cite{snelson-spgp}, and VFE \cite{titsias09}, they all become amenable to our method of PAC-Bayes learning, combining computational benefits of sparse GPs with theoretical guarantees. We carefully account for the different types of parameters (hyperparameters, inducing inputs, observation noise, free-form parameters), as only some of them contribute to the ``penalty term'' in the PAC-Bayes bound. Further, we base GP learning directly on the inverse binary KL divergence \cite{seeger-classification-paper}, and not on looser bounds used previously, such as from Pinsker's inequality (e.g., \cite{nonvacuous-generalization-bounds-snn}).

We demonstrate our GP learning method on regression tasks, whereas PAC-Bayes bounds have so far mostly been used in a classification setting. A PAC-Bayesian bound for regression with potentially unbounded loss function was developed in \cite{pac-bayesian-theory-meets-bayesian-inference}, it requires a sub-Gaussian assumption w.r.t.\ the (unknown) data distribution, see also \cite{excess-risk-bounds-using-vi-in-gaussian-models}. To remain distribution-free as in the usual PAC setting, we employ and investigate a generic \emph{bounded} loss function for regression.

We evaluate our learning method on several datasets and compare its performance to state-of-the-art GP methods \cite{rasmussen-williams-book,snelson-spgp,titsias09} in Sect.\ \ref{experiments-section}. Our learning objective exhibits robust optimization behaviour with the same scaling to large datasets as the other GP methods. We find that our method yields significantly better risk bounds, often by a factor of more than two, and that only for our approach the guarantee improves with the number of inducing points.

\section{General PAC-Bayesian Framework}\label{framework-section}
\subsection{Risk functions}\label{setup-section}
We consider the standard supervised learning setting \cite{understanding-machine-learning-book} where a set $S$ of $N$ training examples $(x_i,y_i)\in X\times Y$ ($i=1,\ldots,N$) is used to learn in a hypothesis space $\mathcal H\subseteq Y^X$, a subset of the space of functions $X\to Y$. We allow learning algorithms that output a distribution $Q$ over hypotheses $h\in\mathcal H$, rather than a single hypothesis $h$, which is the case for GPs we consider later on.

To quantify how well a hypothesis $h$ performs, we assume a bounded loss function $\ell:Y\times Y\to[0,1]$ to be given, w.l.o.g.\ scaled to the interval $[0,1]$. $\ell(y_*,\widehat y)$ measures how well the prediction $\widehat y=h(x_*)$ approximates the actual output $y_*$ at an input $x_*$. The \emph{empirical risk} $R_S(h)$ of a hypothesis is then defined as the average training loss $R_S(h):=\frac{1}{N}\sum_{i=1}^{N} \ell(y_i,h(x_i))$. As in the usual PAC framework, we assume an (unknown) underlying distribution $\mu=\mu(x,y)$ on the set $X\times Y$ of examples, and define the \emph{(true) risk} as $R(h):=\int d\mu(x,y) \ell(y,h(x))$. We will later assume that the training set $S$ consists of $N$ independent draws from $\mu$ and study how close $R_S$ is to its mean $R$ \cite{understanding-machine-learning-book}. To quantify the performance of stochastic learning algorithms, that output a distribution $Q$ over hypotheses, we define the empirical and true risks by a slight abuse of notation as \cite{mcallester-pac-bayesian-model-averaging}:
\begin{align}
R_S(Q):=&\mathbb E_{h\sim Q}\big[ R_S(h)\big] = \frac{1}{N}\sum_{i=1}^N\mathbb E_{h\sim Q}\big[\ell\big(y_i,h(x_i)\big)\big],   \label{stoch-empirical-risk}\\
R(Q):=&\mathbb E_{h\sim Q}\big[ R(h)\big] = \mathbb E_{(x_*,y_*)\sim\mu}\,\mathbb E_{h\sim Q}\big[\ell\big(y_*,h(x_*)\big)\big].    \label{stoch-true-risk}
\end{align}
These are the average losses, also termed \emph{Gibbs risks}, on the training and true distributions, respectively, where the hypothesis $h$ is sampled according to $Q$ before prediction.

In the following, we focus on the regression case, where $Y\subseteq \mathbb R$ is the set of reals.
An exemplary loss function in this case is $\ell(y_*,\widehat y):=\ii_{\widehat y\notin[r_-(y_*),r_+(y_*)]}$, where the functions $r_\pm$ specify an interval outside of which a prediction $\widehat y$ is deemed insufficient; similar to $\varepsilon$-support vector regression \cite{vapnik-the-nature-of-statistical-learning-theory}, we use $r_\pm(y_*):=y_*\pm\varepsilon$, with a desired accuracy goal $\varepsilon>0$ specified before learning (see Sect.\ \ref{experiments-section}).
In any case, the expectations over $h\sim Q$ in (\ref{stoch-empirical-risk})--(\ref{stoch-true-risk}) reduce to one-dimensional integrals as $h(x_*)$ is a real-valued random variable at each $x_*$. See App.\ \ref{derivatives-Phi-appendix}, where we also explore other loss functions.

Instead of the stochastic predictor $h(x_*)$ with $h\sim Q$, one is often interested in the deterministic \emph{Bayes predictor} $\mathbb E_{h\sim Q}[h(x_*)]$ \cite{seeger-classification-paper}; for GP regression, this simply equals the predictive mean $\widehat m(x_*)$ at $x_*$. The corresponding \emph{Bayes risk} is defined by $R_{\rm Bay}(Q):=\mathbb E_{(x_*,y_*)\sim\mu}[\ell(y_*,\mathbb E_{h\sim Q}[h(x_*)])]$. While PAC-Bayesian theorems do not directly give a bound on $R_{\rm Bay}(Q)$ but only on $R(Q)$, it is easy to see that $R_{\rm Bay}(Q)\leq 2R(Q)$ if $\ell(y_*,\widehat y)$ is quasi-convex in $\widehat y$, as in the examples above, and the distribution of $\widehat y=h(x_*)$ is symmetric around its mean (e.g., Gaussian) \cite{seeger-classification-paper}. An upper bound $B(Q)$ on $R(Q)$ below $1/2$ thus implies a nontrivial bound on $R_{\rm Bay}(Q)\leq 2B(Q)<1$.

\subsection{PAC-Bayesian generalization bounds}\label{PAC-Bayesian-thm-section}
In this paper we aim to learn a GP $Q$ by minimizing suitable risk bounds. Due to the probabilistic nature of GPs, we employ generalization bounds for stochastic predictors, which were previously observed to yield stronger guarantees than those for deterministic predictors \cite{seeger-classification-paper,tighter-pac-bayes-bounds-svm,nonvacuous-generalization-bounds-snn}. The most important results in this direction are the so-called ``PAC-Bayesian bounds'', originating from \cite{mcallester-pac-bayesian-model-averaging,asdf2} and developed in various directions \cite{seeger-classification-paper,maurer-note-pac-bayes,catoni-thermodyn-of-ml,pac-bayes-learning-of-linear-classifiers,pac-bayesian-bounds-based-on-the-renyi-divergence,pac-bayesian-theory-meets-bayesian-inference}.

The PAC-Bayesian theorem (Theorem \ref{pac-bayes-theorem}) gives a probabilistic upper bound (generalization guarantee) on the true risk $R(Q)$ of a stochastic predictor $Q$ in terms of its empirical risk $R_S(Q)$ on a training set $S$. It requires to fix a distribution $P$ on the hypothesis space $\mathcal H$ \emph{before} seeing the training set $S$, and applies to the true risk $R(Q)$ of \emph{any} distribution $Q$ on $\mathcal H$\footnote{We follow common usage and call $P$ and $Q$ ``prior'' and ``posterior'' distributions in the PAC-Bayesian setting, although their meaning is somewhat different from priors and posteriors in Bayesian probability theory.}. The bound contains a term that can be interpreted as complexity of the hypothesis distribution $Q$, namely the Kullback-Leibler (KL) divergence $KL(Q\|P):=\int dh\,Q(h)\ln\frac{Q(h)}{P(h)}$, which takes values in $[0,+\infty]$. The bound also contains the binary KL-divergence $kl(q\|p):=q\ln\frac{q}{p}+(1-q)\ln\frac{1-q}{1-p}$, defined for $q,p\in[0,1]$, or more precisely its (upper) inverse $\klinv$ w.r.t.\ the second argument (for $q\in[0,1]$, $\varepsilon\in[0,\infty]$):
\begin{align}\label{define-binary-kl}
\klinv(q, \varepsilon):=\max\{p\in[0,1]\,:\,\kl{q}{p}\leq\varepsilon\},
\end{align}
which equals the unique $p\in[q,1]$ satisfying $\kl{q}{p}=\varepsilon$. While $\klinv$ has no closed-form expression, we refer to App.\ \ref{klinv-derivatives-appendix} for an illustration and more details, including its derivatives for optimization.
\begin{theorem}[PAC-Bayesian theorem \cite{mcallester-pac-bayesian-model-averaging,seeger-classification-paper,maurer-note-pac-bayes}]\label{pac-bayes-theorem}
For any $[0,1]$-valued loss function $\ell$, for any distribution $\mu$, for any $N\in\mathbb N$, for any distribution $P$ on a hypothesis set $\mathcal H$, and for any $\delta\in(0,1]$, the following holds with probability at least $1-\delta$ over the training set $S\sim\mu^N$:
\begin{align}\label{pac-bayes-bound-eq}
\forall Q:\quad R(Q)\,\leq\,\klinv\left(R_S(Q),\frac{\KL{Q}{P}+\ln\frac{2\sqrt{N}}{\delta}}{N}\right).
\end{align}
\end{theorem}The RHS of (\ref{pac-bayes-bound-eq}) can be upper bounded by $R_S(Q)+\sqrt{\big(\KL{Q}{P}+\ln\frac{2\sqrt{N}}{\delta})\big)/(2N)}$, which gives a useful intuition about the involved terms, but can exceed $1$ and thereby yield a trivial statement. Note that the full PAC-Bayes theorem \cite{maurer-note-pac-bayes} gives a simultaneous lower bound on $R(Q)$,
which is however not relevant here as we are going to \emph{minimize} the upper risk bound. Further refinements of the bound
are possible (e.g., \cite{maurer-note-pac-bayes}),
but as they improve over Theorem \ref{pac-bayes-theorem} only in small regimes \cite{catoni-thermodyn-of-ml,pac-bayes-learning-of-linear-classifiers,pac-bayesian-bounds-based-on-the-renyi-divergence}, often despite adjustable parameters, we will stick with the parameter-free bound (\ref{pac-bayes-bound-eq}).

We want to consider a family  of prior distributions $P^\theta$ parametrized by $\theta\in\Theta$, e.g.\ in GP hyperparameter training \cite{rasmussen-williams-book}. If this family is countable, one can generalize the above analysis by fixing some probability distribtion $p_\theta$ on $\Theta$ and defining the mixture prior $P:=\sum_\theta p_\theta P^\theta$; when $\Theta$ is a finite set, the uniform distribution $p_\theta=1/|\Theta|$ is a canonical choice. Using the fact that $\KL{Q}{P}\leq \KL{Q}{P^\theta} +\ln\frac{1}{p_\theta}$ holds for each $\theta\in\Theta$ (App.\ \ref{KL-theta-appendix}), Theorem \ref{pac-bayes-theorem} yields that, with probability at least $1-\delta$ over $S\sim\mu^N$,
\begin{align}\label{pac-bayes-union-bound-eq}
\forall\theta\in\Theta~\forall Q:\quad R(Q)\leq \klinv\left(R_S(Q),\frac{\KL{Q}{P^\theta}+\ln\frac{1}{p_\theta}+\ln\frac{2\sqrt{N}}{\delta}}{N}\right)~=:~B(Q).
\end{align}
The bound (\ref{pac-bayes-union-bound-eq}) holds simultaneously for all $P^\theta$ and all $Q$\footnote{The same result can be derived from (\ref{pac-bayes-bound-eq}) via a union bound argument (see Appendix \ref{KL-theta-appendix}).}. One can thus optimize over both $\theta$ and $Q$ to obtain the best generalization guarantee, with confidence at least $1-\delta$. We use $B(Q)$ for our training method below, but we will also compare to training with the suboptimal upper bound $B_{\rm Pin}(Q):=R_S(Q)+\sqrt{\big(KL(Q\|P^\theta)+\ln\frac{1}{p_\theta}+\ln\frac{2\sqrt{N}}{\delta}\big)/(2N)}\geq B(Q)$ as was done previously \cite{nonvacuous-generalization-bounds-snn}. The PAC-Bayesian bound depends only weakly on the confidence parameter $\delta$, which enters logarithmically and is suppressed by the sample size $N$. When the hyperparameter set $\Theta$ is not too large (i.e.\ $ \ln \frac{1}{p_\theta}$ is small compared to $N$), the main contribution to the \emph{penalty term} in the second argument of $kl^{-1}$ comes from $\frac{1}{N}KL(Q\|P^\theta)$, which must be $\ll1$ for a good generalization statement (see Sect.\ \ref{experiments-section}).

\section{PAC-Bayesian learning of GPs}\label{learning-GPs-section}


\subsection{Learning full GPs}\label{full-GP-section}
GP modelling is usually presented as a Bayesian method \cite{rasmussen-williams-book}, in which the prior $P(f)=\mathcal{GP}(f{\mid}m(x),K(x,x'))$ is specified by a positive definite kernel $K:X\times X\to\mathbb R$ and a mean function $m:X\to\mathbb R$ on the input set $X$.
In ordinary GP regression, the learned distribution $Q$ is then chosen as the Bayesian posterior coming from the assumption that the training outputs
$\trainoutputs:=(y_i)_{i=1}^N\in\mathbb R^N$ are noisy versions of $f_N=(f(x_1),\ldots,f(x_N))$ with i.i.d.\ Gaussian likelihood $\trainoutputs|f_N\sim\mathcal N(\trainoutputs|f_N,\sigma_n^2\ii)$. Under this assumption, $Q$ is again a GP \cite{rasmussen-williams-book}:
\begin{equation}\label{usual-GPR-posterior}
\begin{split}
Q(f)\,=\,\mathcal{GP}\big(f \mid \>&m(x)+k_N(x)(K_{NN}+\sigma_n^2\ii)^{-1}(\trainoutputs-m_N),\\
&\>K(x,x')-k_N(x)(K_{NN}+\sigma_n^2\ii)^{-1}k_N(x')^T\big),
\end{split}
\end{equation}
with $K_{NN}=(K(x_i,x_j))_{i,j=1}^N$, $k_N(x)=(K(x,x_1),\ldots,K(x,x_N))$, $m_N=(m(x_1),\ldots,m(x_N))$. Eq.\ (\ref{usual-GPR-posterior}) is employed to make (stochastic) predictions for $f(x_*)$ on new inputs $x_*\in X$. In our approach below, we do not require any Bayesian rationale behind $Q$ but merely use its form, parametrized by $\sigma_n^2$, as an optimization ansatz within the PAC-Bayesian theorem.

Importantly, for any full GP prior $P$ and its corresponding posterior $Q$ from (\ref{usual-GPR-posterior}), the KL-divergence $KL(Q\|P)$ in Theorem \ref{pac-bayes-theorem} and Eq.\ (\ref{pac-bayes-union-bound-eq}) can be evaluated on \emph{finite} ($N$-)dimensional matrices. 
This allows us to evaluate the PAC-Bayesian bound and in turn to learn GPs by optimizing it.
 More precisely, one can easily verify that $P$ and $Q$ have the \emph{same} conditional distribution $P(f|f_N)=Q(f|f_N)$\footnote{In fact, direct computation \cite{rasmussen-williams-book} gives $P(f|f_N)=\mathcal{GP}\big(f|m(x)+k_N(x)K_{NN}^{-1}(f_N-m_N),K(x,x')-k_N(x)K_{NN}^{-1}k_N(x')^T\big)=Q(f|f_N)$. 
 	Remarkably, $Q(f|f_N)$ does \emph{not} depend on $\trainoutputs$ nor on $\sigma_n$, even though $Q(f)$ from (\ref{usual-GPR-posterior}) does. Intuitively this is because, for the above likelihood, $f$ is independent of $\trainoutputs$ given $f_N$.}, so that
\begin{align}
\KL{Q}{P} &= \KL{Q(f_N)Q(f \mid f_N)}{P(f_N)P(f \mid f_N)}
\
=\KL{Q(f_N)}{P(f_N)}   \label{make-KL-finite-dim}  \\
&=\frac{1}{2}\ln\det\big[K_{NN}+\sigma_n^2\ii\big]-\frac{N}{2}\ln\sigma_n^2-\frac{1}{2}{\rm tr}\big[K_{NN}(K_{NN}+\sigma_n^2\ii)^{-1}\big]\label{non-sparse-KL-formula}\\
&\ \ +\frac{1}{2}(\trainoutputs-m_N)^T(K_{NN}+\sigma_n^2\ii)^{-1}K_{NN}(K_{NN}+\sigma_n^2\ii)^{-1}(\trainoutputs-m_N),\nonumber
\end{align}
where in the last step we used the well-known formula \cite{kullback-leibler-paper} for the KL divergence between normal distributions $P(f_N)=\mathcal N(f_N \mid m_N,K_{NN})$ and $Q(f_N)=\mathcal N\big(f_N \mid m_N+K_{NN}(K_{NN}+\sigma_n^2\ii)^{-1}(\trainoutputs-m_N),K_{NN}-K_{NN}(K_{NN}+\sigma_n^2\ii)^{-1}K_{NN}\big)$, and simplified a bit (see also App.\ \ref{other-objective-functions-section}).

To learn a full GP means to select ``good'' values for the \emph{hyperparameters} $\theta$, which parametrize a family of GP priors $P^\theta={\mathcal{GP}}(f{\mid}m^\theta(x),K^\theta(x,x'))$, and for the noise level $\sigma_n$ \cite{rasmussen-williams-book}. Those values are afterwards used to make predictions with the corresponding posterior $Q^{\theta,\sigma_n}$ from (\ref{usual-GPR-posterior}). In our experiments (Sect.\ \ref{experiments-section}) we will use the squared exponential (SE) kernel on $X=\mathbb R^d$, $K^\theta(x,x')=\sigma_s^2\exp[-\frac{1}{2}\sum_{i=1}^d\frac{(x_i-x'_i)^2}{l_i^2}]$, where
$\sigma_s^2$ is the signal variance, $l_i$ are the lengthscales, and we set the mean function to zero.
The hyperparameters are $\theta\equiv(l_1^2,\ldots,l_d^2,\sigma_s^2)$ (SE-ARD kernel \cite{rasmussen-williams-book}), or $\theta\equiv(l^2,\sigma_s^2)$ if we take all lengthscales $l_1=\ldots=l_d\equiv l$ to be equal (non-ARD).

The basic idea of our method, which we call ``PAC-GP'' is now to learn the parameters\footnote{Contrary to the usual GP viewpoint \cite{rasmussen-williams-book}, $\sigma_n$ is \emph{not} a hyperparameter in our method since the prior $P^\theta$ does not depend on $\sigma_n$. Thus, $\sigma_n$ does also not contribute to the ``penalty term'' $\ln|\Theta|$. $\sigma_n$ is merely a free parameter in the posterior distribution $Q^{\theta,\sigma_n}$. By (\ref{non-sparse-KL-formula}), $KL(Q^{\theta,\sigma_n}\|P^\theta)\to\infty$ as $\sigma_n\to0$, so we need this parameter $\sigma_n^2>0$ because otherwise $KL=\infty$ and the bound as well as the optimization objective would become trivial. Although the parameter $\sigma_n^2$ is originally motivated by a Gaussian observation noise assumption, the aim here is merely to parameterize the posterior in some way while maintaining computational tractability; cf.\ also Sect.\ \ref{sparse-GP-subsection}.} $\theta$ and $\sigma_n$ by minimizing the upper bound $B(Q^{\theta,\sigma_n})$ from Eq.\ (\ref{pac-bayes-union-bound-eq}), therefore selecting the GP predictor $Q^{\theta,\sigma_n}$ with the best generalization performance guarantee within the scope of the PAC-Bayesian bound. Note that all involved terms $R_S(Q^{\theta,\sigma_n})$ (App.\ \ref{derivatives-Phi-appendix}) and $KL(Q^{\theta,\sigma_n}\|P^{\theta})$ from (\ref{non-sparse-KL-formula}) as well as their derivatives (App.\ \ref{klinv-derivatives-appendix}) can be computed effectively, so we can use gradient-based optimization.

The only remaining issue is that the learned prior hyperparameters $\theta$ have to come from a discrete set $\Theta$ that must be specified before seeing the training set $S$ (Sect.\ \ref{PAC-Bayesian-thm-section}). To achieve this, we first minimize the RHS of Eq.\ (\ref{pac-bayes-union-bound-eq}) over $\theta$ and $\sigma_n^2$ in a gradient-based manner, and thereafter discretize each of the components of $\ln\theta$ to the closest point in the equispaced $(G+1)$-element set $\{-L,-L+\frac{2L}{G},\ldots,+L\}$; thus, when $T$ denotes the number of components of $\theta$, the penalty term to be used in the optimization objective (\ref{pac-bayes-union-bound-eq}) is $\ln\frac{1}{p_\theta}=\ln|\Theta|=T\ln(G+1)$. The SE-ARD kernel has $T=d+1$, while the standard SE kernel has $T=2$ parameters. In our experiments we round each component of $\ln\theta$ to two decimal digits in the range $[-6,+6]$, i.e.\ $L=6$, $G=1200$. We found that this discretization has virtually no effect on the predictions of $Q^{\theta,\sigma_n}$, and that coarser rounding (i.e.\ smaller $|\Theta|$) does not significantly improve the bound (\ref{pac-bayes-union-bound-eq}) (via its smaller penalty term $\ln|\Theta|$) nor the optimization (via its higher sensitivity to $Q$); see App.\ \ref{sec:exper-depend-upper}.


\subsection{Learning sparse GPs}\label{sparse-GP-subsection}
Despite the fact that, with confidence $1-\delta$, the bound in (\ref{pac-bayes-union-bound-eq}) holds for \emph{any} $P_\theta$ from the prior GP family and for \emph{any} distribution $Q$, we optimized in Sect.\ \ref{full-GP-section} the upper bound merely over the parameters $\theta,\sigma_n$ after substituting $P^{\theta}$ and the corresponding $Q^{\theta,\sigma_n}$ from (\ref{usual-GPR-posterior}). We are limited by the need to compute $KL(Q\|P)$ effectively, for which we relied on the property $Q(f\mid f_N)=P(f\mid f_N)$ and the Gaussianity of $P(f_N)$ and $Q(f_N)$, cf.\ (\ref{make-KL-finite-dim}). Building on this two requirements, we now construct more general pairs $P,Q$ of GPs with effectively computable $KL(Q\|P)$, so that our learning method becomes more widely applicable, including sparse GP methods.


Instead of the points $x_1,\ldots,x_N$ associated with the training set $S$ as in Sect.\ \ref{full-GP-section}, one may choose from the input space any number $M$ of points $Z=\{z_1,\ldots,z_M\}\subseteq X$, often called \emph{inducing inputs}, and any Gaussian distribution $Q(f_M)=\mathcal N(f_M \mid a_M,B_{MM})$ on function values $f_M:=(f(z_1),\ldots,f(z_M))$, with any $a_M\in\mathbb R^M$ and positive semidefinite matrix $B_{MM}\in\mathbb R^{M\times M}$. The distribution $Q$ on $f_M$ can be extended to all function values at all inputs $X$ using the conditional $Q(f \mid f_M)=P(f \mid f_M)$ from the prior $P$ (see Sect.\ \ref{full-GP-section}). This yields the following predictive GP:
\begin{align}\label{posterior-general-sparse-GP}
Q(f)\,=\,\mathcal{GP}\big(f \mid \>&m(x)+k_M(x)K_{MM}^{-1}(a_M-m_M),\\
&\>K(x,x')-k_M(x)K_{MM}^{-1}[K_{MM}-B_{MM}]K_{MM}^{-1}k_M(x')^T\big),\nonumber
\end{align}
where $K_{MM}:=(K(z_i,z_j))_{i,j=1}^M$, $k_{M}(x):=(K(x,z_1),\ldots,K(x,z_M))$, and $m_M:=(m(z_1),\ldots,m(z_M))$. This form of $Q$ includes several approximate posteriors from Bayesian inference that have been used in the literature \cite{rasmussen-williams-book,seeger-classification-paper,titsias09,scalable-variational-gaussian-process-classification}, even for noise models other than the Gaussian one used to motivate the $Q$ from Sect.\ \ref{full-GP-section}. Analogous reasoning as in (\ref{make-KL-finite-dim}) now gives \cite{seeger-classification-paper,titsias09,KL-divergence-between-stochastic-processes}:
\begin{align}
\KL{Q}{P}=\KL{Q(f_M)}{P(f_M)}=&-\frac{1}{2}\ln\det\big[B_{MM}K_{MM}^{-1}\big]+\frac{1}{2}{\rm tr}\big[B_{MM}K_{MM}^{-1}\big]-\frac{M}{2}  \nonumber\\
&+\frac{1}{2}(a_M-m_M)^T K_{MM}^{-1} (a_M-m_M).   \label{KL-for-sparse-GP}
\end{align}
One can thus effectively optimize in (\ref{pac-bayes-union-bound-eq}) the prior $P^{\theta}$ and the posterior distribution $Q^{\theta,\{z_i\},a_M,B_{MM}}$ by varying the number $M$ and locations $z_1,\ldots,z_M$ of inducing inputs and the parameters $a_M$ and $B_{MM}$, along with the hyperparameters $\theta$.
Optimization can in this framework be organized such that it consumes time $O(NM^2+M^3)$ per gradient step and memory $O(NM+M^2)$ as opposed to $O(N^3)$ and $O(N^2)$ for the full GP of Sect.\ \ref{full-GP-section}.
This is a big saving when $M\ll N$ and justifies the name ``sparse GP'' \cite{rasmussen-williams-book,unifying-view-sparse-gp-approx}.

Some popular sparse-GP methods \cite{unifying-view-sparse-gp-approx} are special cases of the above form, by prescribing certain $a_M$ and $B_{MM}$ depending on the training set $S$, so that only the inducing inputs $z_1,\ldots,z_M$ and a few parameters such as $\sigma_n^2$ are left free:
\begin{align}
a_M= K_{MM}Q_{MM}^{-1}K_{MN}(\alpha\Lambda+\sigma_n^2\ii)^{-1}\trainoutputs,\quad B_{MM}=&K_{MM}Q_{MM}^{-1}K_{MM}  \label{BMM-sparse-eqn},
\end{align}
where $Q_{MM}=K_{MM}+K_{MN}(\alpha\Lambda+\sigma_n^2\ii)^{-1}K_{NM}$ with $K_{MN}:=(K(z_i,x_j))_{i,j=1}^{M,N}$, $K_{NM}=K_{MN}^T$, and $\Lambda=\diag(\lambda_1,\ldots,\lambda_N)$ is a diagonal $N\times N$-matrix with entries $\lambda_i=K(x_i,x_i)-k_M(x_i)K_{MM}^{-1}k_M(x_i)^T$. Setting $\alpha=1$ corresponds to the FITC approximation \cite{snelson-spgp}, whereas $\alpha=0$ is the VFE and DTC method \cite{titsias09,seeger-dtc} (see App.\ \ref{other-objective-functions-section} for their training objectives); one can also linearly interpolate between both choices with $\alpha\geq0$ \cite{bui-yan-turner-unifying}. Another form of sparse GPs where the latent function values $f_M$ are fixed and not marginalized over, corresponds to $B_{MM}=0$, which however gives diverging $\KL{Q}{P}=\infty$ via (\ref{KL-for-sparse-GP}) and therefore trivial  bounds in (\ref{pac-bayes-bound-eq})--(\ref{pac-bayes-union-bound-eq}).




Our learning method for sparse GPs (``PAC-SGP'') follows now similar steps as in Sect.\ \ref{full-GP-section}: One has to include a penality $\ln\frac{1}{p_\theta}=\ln|\Theta|$ for the prior hyperparameters $\theta$, which are to be discretized into the set $\Theta$ after the optimization of (\ref{pac-bayes-union-bound-eq}). Note, $\theta$ contains the prior hyperparameters only and not the inducing points $z_1,\ldots,z_M$ nor $a_M$, $B_{MM}$, $\sigma_n$, or $\alpha$ from (\ref{BMM-sparse-eqn}); all these quantities can be optimized over simultaneously with $\theta$, but do not need to be discretized. The number $M$ of inducing inputs can also be varied, which determines the required computational effort, and all optimizations can be both discrete \cite{seeger-dtc} or continuous \cite{snelson-spgp,titsias09}. When optimizing over positive $B_{MM}$, the parametrization $B_{MM}=LL^T$ with a lower triangular matrix $L\in\mathbb R^{M\times M}$ can be used \cite{scalable-variational-gaussian-process-classification}. For the experiments below we always employ the FITC parametrization (fixed $\alpha=1$) in our proposed PAC-SGP method, i.e.\ our optimization parameters are $\sigma_n^2$ and $\{z_i\}$ besides the length scale hyperparameters $\theta$.




\section[Experiments]{Experiments\protect\footnote{Python code (building on GPflow \cite{gpflow} and TensorFlow \cite{tensorflow}) implementing our method is available at \url{https://github.com/boschresearch/PAC\_GP}.}}
\label{experiments-section}

We now illustrate our learning method and compare it with other GP methods on various regression tasks.
In contrast to prior work \cite{nonvacuous-generalization-bounds-snn}, we found the gradient-based training with the objective \eqref{pac-bayes-union-bound-eq} to be robust enough, such that no pretraining with conventional objectives (such as from App.\ \ref{other-objective-functions-section}) is necessary.
We set $\delta=0.01$ throughout \cite{seeger-classification-paper,nonvacuous-generalization-bounds-snn}, cf.\ Sect.\ \ref{PAC-Bayesian-thm-section}, and use (unless specified otherwise) the generic bounded loss function $\ell(y,\widehat y)=\ii_{\widehat y\notin[y-\varepsilon,y+\varepsilon]}$ for regression, with accuracy goal $\varepsilon>0$ as specified below.

We evaluate the following methods:
\emph{(a) PAC-GP:} Our proposed method (cf.\,Sect.\,\ref{full-GP-section}) with the training objective $B(Q)$ \eqref{pac-bayes-union-bound-eq} (kl-PAC-GP) and for comparison with the looser training objective $B_{\rm Pin}$ (sqrt-PAC-GP) (see below \eqref{pac-bayes-union-bound-eq}, similar to e.g.\ \cite{nonvacuous-generalization-bounds-snn});
\emph{(b) PAC-SGP:} Our sparse GP method (Sect.\ \ref{sparse-GP-subsection}), again with objectives $B(Q)$ (kl-PAC-SGP) and $B_{\rm Pin}(Q)$ (sqrt-PAC-SGP), respectively;
\emph{(c) full-GP:} The ordinary full GP for regression \cite{rasmussen-williams-book};
\emph{(d) VFE:} Titsias' sparse GP \cite{titsias09};
\emph{(e) FITC:} Snelson-Ghahramani's sparse GP \cite{snelson-spgp}.
Note that full-GP, VFE, and FITC as well as sqrt-PAC-GP and sqrt-PAC-SGP are \emph{trained} on other objectives (see App.\ \ref{other-objective-functions-section}), and we will \emph{evaluate} the upper bound \eqref{pac-bayes-union-bound-eq} on their generalization performance by evaluating $\KL{Q}{P}$ via \eqref{non-sparse-KL-formula} or \eqref{KL-for-sparse-GP}.
To obtain finite generalization bounds, we discretize $\theta$ for all methods at the end of training as in Sect.\ \ref{full-GP-section} and use the appropriate $\ln\frac{1}{p_\theta}=\ln|\Theta|$ in (\ref{pac-bayes-union-bound-eq}).

\begin{figure}[t]
\includegraphics[width=\textwidth]{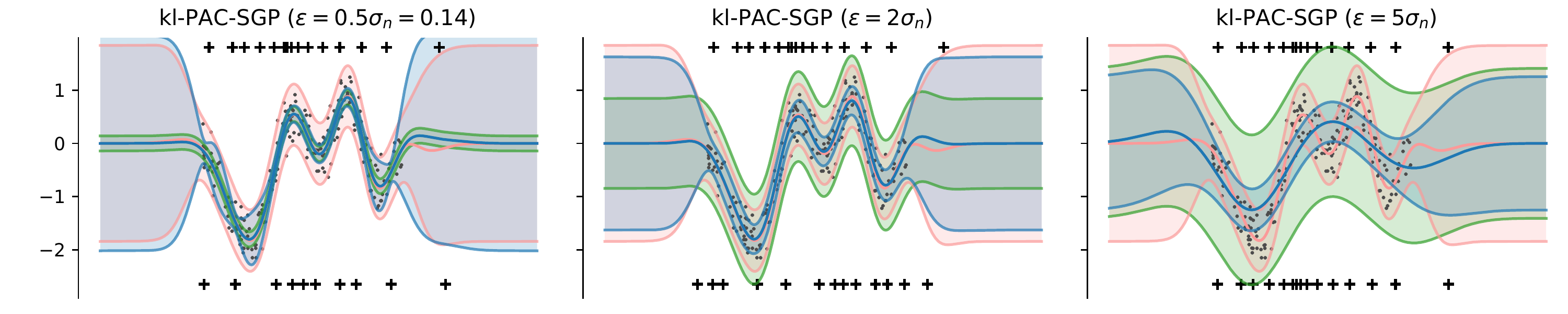}
\caption{\textbf{Predictive distributions.} The predictive distributions (mean $\pm 2\sigma$ as shaded area) of our kl-PAC-SGP (blue) are shown for various choices of $\varepsilon$ together with the full-GP's prediction (red). (Note that by Eqs.\ (\ref{posterior-general-sparse-GP},\ref{BMM-sparse-eqn}), kl-PAC-SGP's predictive variance does \emph{not} include additive $\sigma_n^2$, whereas full-GP's does \cite{rasmussen-williams-book}.) The shaded green area visualizes an $\varepsilon$-band, centered around the kl-PAC-SGP's predictive mean; datapoints (black dots) inside this band do not contribute to the risk $R_S(Q)$. Crosses above/below the plots indicate the inducing point positions ($M=15$) before/after training.\label{snelson-fig}}
\end{figure}

\paragraph{(a) Predictive distribution.} To get a first intuition, we illustrate in Fig.\ \ref{snelson-fig} the effect of varying $\varepsilon$ in the loss function on the predictive distribution of our sparse PAC-SGP.
The accuracy goal $\varepsilon$ defines a band around the predictive mean within which data-points do not contribute to the empirical risk $R_S(Q)$. We thus chose the accuracy goal $\varepsilon$ relative to the observation noise $\sigma_n$ obtained from an ordinary full-GP. Results are presented on the 1D toy dataset\footnote{snelson: dimensions 200 $\times$ 1, available at \url{www.gatsby.ucl.ac.uk/~snelson}.} from the original FITC \cite{snelson-spgp} and VFE \cite{titsias09} publications (for a comparison to the predictive distributions of FITC and VFE see App.\ \ref{sec:exper-comp-sparse}, which also contains an illustration that our kl-PAC-SGP avoids FITC's known overfitting on pathological datasets.). Here and below, we optimize the hyperparameters in each experiment anew.

We find that for large $\varepsilon$ (right plot) the predictive distribution (blue) becomes smoother: Due to the wider $\varepsilon$-band (green), the PAC-SGP does not need to adapt much to the data for the $\varepsilon$-band to contain many data points.
Hence the predictive distribution can remain closer to the prior, which reduces the $\text{KL}$-term in the objective \eqref{pac-bayes-union-bound-eq}.
For the same reason, the inducing points need not adapt much compared to their initial positions for large $\varepsilon$.
For smaller $\varepsilon$, the PAC-SGP adapts more to the data, whereas for very small $\varepsilon$ (left plot), it is anyhow not possible to place many data points within the narrow $\varepsilon$-band, so the predictive distribution can again be closer to the prior (compare e.g.\ in the first and second plots the blue curves near the rightmost datapoints) for a smaller $\text{KL}$-term.
In particular, the KL-divergence (divided by number of training points) for the three settings in Fig.\ref{snelson-fig} are: 0.097 (left), 0.109 (middle), and 0.031 (right).

\paragraph{(b) Full-GP experiments -- dependence on the accuracy goal $\varepsilon$.}

\begin{figure}
\includegraphics[width=\textwidth]{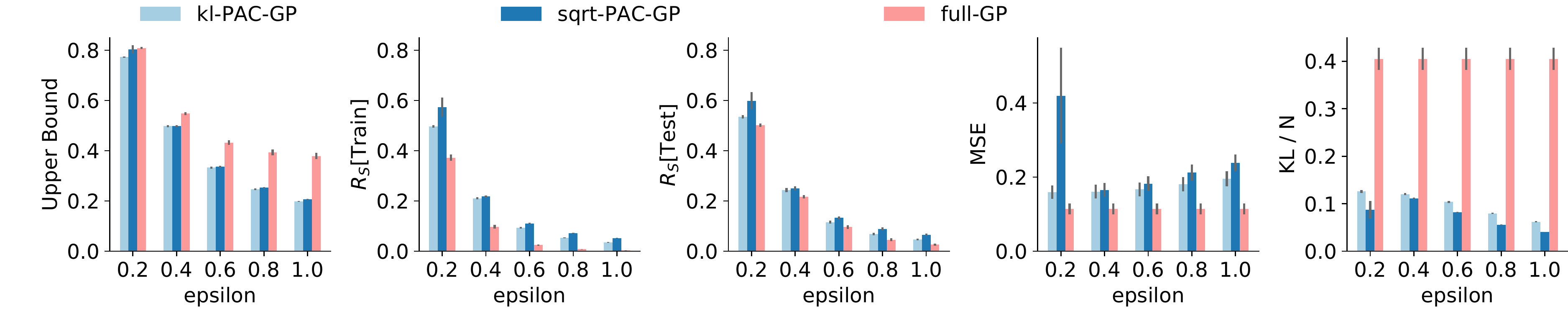}
\caption{\textbf{Dependence on the accuracy goal $\varepsilon$.} For each $\varepsilon$, the plots from left to right show (means as bars, standard errors after ten iterations as grey ticks) the upper bound $B(Q)$ from Eq.\ (\ref{pac-bayes-union-bound-eq}), the Gibbs training risk $R_S(Q)$, the Gibbs test risk as a proxy for the true $R(Q)$, MSE, and $\KL{Q}{P^{\theta}}/N$, after learning $Q$ on the dataset \textit{boston housing} by three different methods: our kl-PAC-GP method from Sect.\ \ref{full-GP-section} with sqrt-PAC-GP and the ordinary full-GP.   \label{boston-fig}  }
\end{figure}

To explore the dependence on the desired accuracy $\varepsilon$ further, we compare in Fig.\ \ref{boston-fig} the ordinary full-GP to our PAC-GPs on the \textit{boston housing dataset}\footnote{boston: dimensions 506 $\times$ 13, available at \url{http://lib.stat.cmu.edu/datasets/boston}}. As pre-processing we normalized all features and the output to mean zero and unit variance, then analysed the impact of the accuracy goal $\varepsilon \in \{0.2, 0.4, 0.6, 0.8, 1.0\}$. We used 80\% of the dataset for training and 20\% for testing, in ten repetitions of the experiment.

Our PAC-GP yields significantly better generalization guarantees for all accuracy goals $\varepsilon$ compared to full-GP, since we are directly optimizing the bound \eqref{pac-bayes-union-bound-eq}. This effect is stronger for large $\varepsilon$, where the KL-term of PAC-GP can decrease as $Q$ may again remain closer to $P$ while keeping the training loss low. 
Although better bounds do not necessarily imply better Gibbs test risk, kl-PAC-GP performs only slightly below the  ordinary full-GP in this regard. 
Moreover, our PAC-GPs exhibit less overfitting than the full-GP, for which the training risks are significantly larger than the test risks (see Table \ref{table-appendix-boston} in App.\ \ref{experimental-data-appendix} for numerical values).
On the other hand, the tighter objective \eqref{pac-bayes-union-bound-eq} in the kl-PAC-GP allows learning a slightly more complex GP $Q$ in terms of the KL-divergence compared to the sqrt-PAC-GP, which results in better test risks and at the same time better guarantees.
This confirms that kl-PAC-GP is always preferable to sqrt-PAC-GP.
However, as any prediction within the full $\pm\varepsilon$-band around the ground truth incurs no risk for our PAC-GPs, their mean squared error (MSE) increases with $\varepsilon$.

The fact that our learned PAC-GPs exhibit higher training and test errors (Gibbs risk and esp.\ MSE) than full-GP can be explained by their \emph{under}fitting in order to hedge against violating Eq.\ (\ref{pac-bayes-union-bound-eq}) (i.e.\ Theorem \ref{pac-bayes-theorem}). This underfitting is evidenced by PAC-GP's significantly less complex learned posterior $Q$ as measured by $KL(Q\|P^\theta)/N$ (Fig.\ \ref{boston-fig}), or similarly (via Eqs.\ (\ref{non-sparse-KL-formula},\ref{KL-for-sparse-GP})), by its larger learned noise variance $\sigma_n^2$ compared to full-GP's (Table \ref{table-appendix-boston} in App.\ \ref{experimental-data-appendix}). It is exactly this stronger regularization of PAC-GP in terms of the $KL$ divergence that leads to its better generalization guarantees.

In the following, we will fix $\varepsilon=0.6$ after pre-processing data as above, to illustrate PAC-GP further. Note however that in a concrete application, $\varepsilon$ should be fixed to a desired accuracy goal using domain knowledge, but before seeing the training set $S$. Alternatively, one can consider a set of $\varepsilon$-values $\varepsilon_1,\ldots,\varepsilon_E$ chosen in advance, at the cost of a term $\ln E$ in addition to $\log\frac{1}{p_\theta}$ in the objective (\ref{pac-bayes-union-bound-eq}).

\paragraph{(c) Sparse-GP experiments -- dependence on number of inducing inputs $M$.}

\begin{figure}
\includegraphics[width=\textwidth]{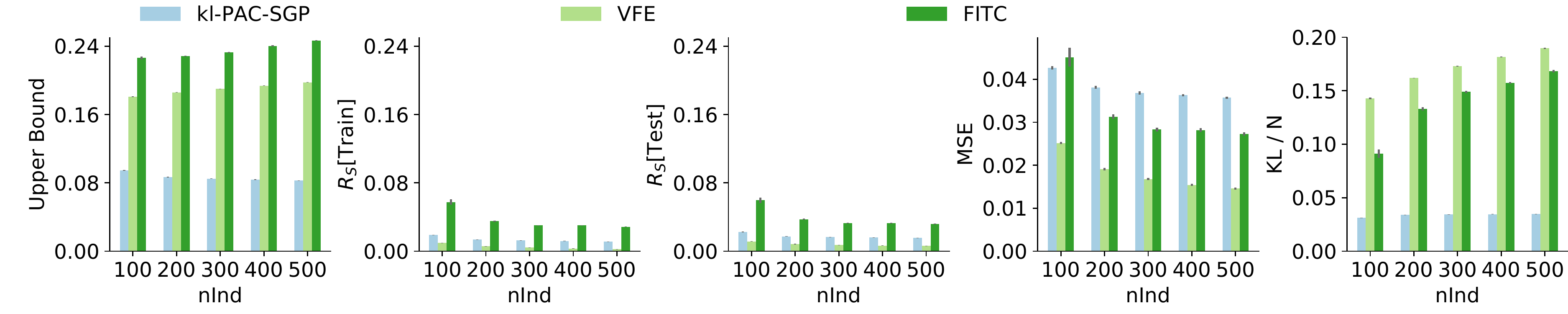}
\includegraphics[width=\textwidth]{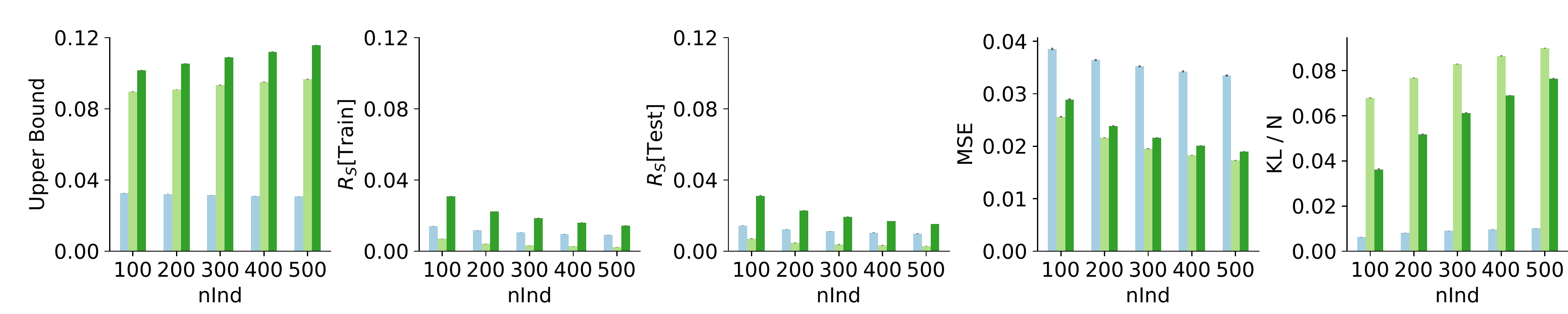}
\includegraphics[width=\textwidth]{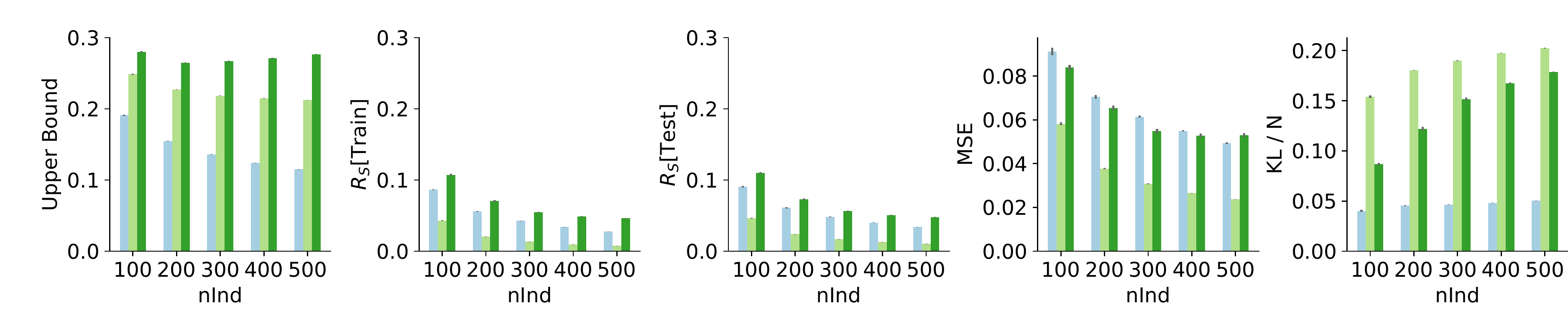}
\caption{\textbf{Dependence on the number of inducing variables.} 
Shown is the average  ($\pm$ standard error over $10$ repetitions) upper bound $B$, Gibbs training risk $R_S$, Gibbs test risk, MSE, and $KL(Q\|P^{\theta})/N$ as a function of the number $M$ of inducing inputs (from left to right). We compare our sparse kl-PAC-SGP  (Sect.\ \ref{sparse-GP-subsection}) with the two popular GP approximations VFE and FITC. Each row corresponds to one dataset: \textit{pol} (top), \textit{sarcos} (middle) and \textit{kin40k} (bottom). kl-PAC-SGP has the best guarantee in all settings (left column), due to a lower model complexity (right column), but this comes at the price of slightly larger test errors.  \label{sparse-fig}  }
\end{figure}

We now examine our sparse PAC-SGP method (Sect.\ \ref{sparse-GP-subsection}) on the three large data sets  \textit{pol, sarcos}, and \textit{kin40k}\footnote{pol: 15,000 $\times$ 26, kin40k: 40,000 $\times$ 8 (both from \url{https://github.com/trungngv/fgp.git});\\sarcos: 48,933 $\times$ 21 (\url{http://www.gaussianprocess.org/gpml/data})}, again using 80\%--20\% train-test splits and ten iterations. 
The results are shown in Fig.\ \ref{sparse-fig}. 
Here, we vary the number of inducing points $M\in\{100,200,300,400,500\}$. For modelling \textit{pol} and \textit{kin40k}, we use the SE-ARD kernel due to its better performance on these datasets, whereas we model \textit{sarcos} without ARD  (cf.\ Table  \ref{table-appendix-sparse-GPs} in App.\ \ref{experimental-data-appendix} for the comparison ARD vs.\ non-ARD).
  The corresponding penalty terms for the three plots are $\frac{1}{N}\ln|\Theta|=0.0160,0.0003,0.0020$ and $\frac{1}{N}\ln\frac{2\sqrt{N}}{\delta}=0.0008, 0.0003, 0.0003$; when compared to ${KL}(Q\|P)/N$ from Fig.\ \ref{sparse-fig}, their contribution is largest for the \textit{pol} dataset.
 
Our kl-PAC-SGP achieves significantly better upper bounds than VFE and FITC, by more than a factor of 3 on \textit{sarcos}, a factor of roughly 2 on \textit{pol}, and a factor between 1.3 and 2 on \textit{kin40k} (Fig.\ \ref{sparse-fig}, cf.\ also Table \ref{table-appendix-sparse-GPs} in App.\ \ref{experimental-data-appendix}).
Also, the PAC-Bayes upper bound is much tighter for kl-PAC-SGP than for VFE or FITC, i.e.\ closer to the Gibbs risk, often by factors exceeding 3.
Our kl-PAC-SGP behaves also more favorably in terms of generalization guarantee when inducing points are added and more complex models are allowed: our upper bound improves substantially with $M$ (\textit{kin40}) or does at least not degrade (\textit{pol} and \textit{sarcos}), as opposed to VFE and FITC, whose complexities $KL/N$ grow substantially with $M$.
   Since very low training risks can already be achieved by a moderate number of inducing points for \textit{pol} and \textit{sarcos}, a growing $KL$ with $M$ deteriorates the upper bound. 
   Regarding the upper bound, the increased flexibility from larger $M$ only pays off for the \textit{kin40k} dataset, whereas the MSE improves with increasing $M$ for all models and datasets.
As above, kl-PAC-SGP is always slightly preferrable to sqrt-PAC-SGP, not only for the upper bound and Gibbs risks as expected but also for MSE (see Table \ref{table-appendix-sparse-GPs} in App.\ \ref{experimental-data-appendix}).

Similarly to the boston dataset, the higher test errors of kl-PAC-SGP compared to VFE and FITC can be explained by underfitting due to the stronger regularization, again shown by lower $KL$ and significantly larger learned $\sigma_n^2$ (by factors of 4--28 compared to VFE), cf.\ Table \ref{table-appendix-sparse-GPs} in App.\ \ref{experimental-data-appendix}. In fact, although our implementation of PAC-SGP employs the FITC parametrization, the PAC-(S)GP optimization is not prone to FITC's well-known overfitting tendency \cite{bauer-understanding-sparse-GP-approximations}, due to the regularization via the $KL$-divergence (see App.\ \ref{sec:exper-comp-sparse}, and in particular Supplementary Figure \ref{fig:objective-scatter}).


To investigate whether the higher test MSE of PAC-GP compared to VFE and FITC (and the full-GP above) is a consequence of the 0-1-loss $\ell(y,\widehat y)=\ii_{\widehat y\notin[y-\varepsilon,y+\varepsilon]}$ used so far, we re-ran the PAC-SGP experiments for $M=500$ inducing inputs with the more distance-sensitive loss function $\ell_{\exp}(y,\widehat y)=1-\exp[-((y-\widehat y)/\varepsilon)^2]$ (Eq.\ (\ref{loss-function-exp})), which is MSE-like for small deviations $|y-\widehat{y}|\lesssim\varepsilon$, i.e.\ $\ell_{\exp}(y,\widehat y)\approx(y-\widehat y)^2/\varepsilon^2$ (Supplementary Figure \ref{loss-functions-plot}). Our results are tabulated in Table \ref{table-appendix-inverted-Gaussian-loss} in App.\ \ref{experimental-data-appendix}. The findings are inconclusive and range from an improvement w.r.t.\ MSE of 25\% (\emph{pol}) over little change (\emph{sarcos}) to a decline of 12\% (\emph{kin40k}), showing that the effect of the loss function is smaller than might have been expected. Nevertheless, generalization guarantees of PAC-SGP remain much better than the ones of the other methods. While the MSE of our PAC-GPs would improve by choosing smaller $\varepsilon$ (e.g., Fig.\ \ref{boston-fig}), this comes at the disadvantage of worse generalization bounds.


We further note that no method shows significant overfitting in Fig.\ \ref{sparse-fig}, in the sense that the differences between test and training Gibbs risks are all rather small, despite the KL-complexity increasing with $M$ for VFE and FITC. This is unlike for \textit{Boston housing} above, and may be due to the much larger training sets here. When comparing VFE and FITC, we observe that VFE consistently outperforms FITC in terms of both MSE as well as generalization guarantee, where VFE's higher KL-complexity is offset by its much lower Gibbs risk. This fortifies the results in \cite{bauer-understanding-sparse-GP-approximations}. We lastly note that, since for our PAC-SGP the obtained guarantees $B$ are much smaller than 1/2, we obtain strong guarantees even on the Bayes risk $R_{\rm Bay}\leq 2B<1$ (Sect.\ \ref{setup-section}).

\section{Conclusion}\label{conclusion-section}
In this paper, we proposed and explored the use of PAC-Bayesian bounds as an optimization objective for GP training. Consequently, we were able to achieve significantly better guarantees on the out-of-sample performance compared to state-of-the-art GP methods, such as VFE or FITC, while maintaining computational scalability. We further found that using the tighter generalization bound $B(Q)$ (\ref{pac-bayes-union-bound-eq}) based on the inverse binary kl-divergence leads to an increase in the performance on all metrics compared to a looser bound $B_{\rm Pin}$ as employed in previous works (e.g.\ \cite{nonvacuous-generalization-bounds-snn}).

Despite the much better generalization guarantees obtained by our method, it often yields worse test error, in particular test MSE, than standard GP regression methods; this largely persists even when using more distance-sensitive loss functions than the 0-1-loss. The underlying reason could be that all loss functions considered in this work were bounded, as necessitated by our desire to provide generalization guarantees irrespective of the true data distribution. While rigorous PAC-Bayesian bounds exist for MSE-like unbounded loss functions under special assumptions on the data distribution \cite{pac-bayesian-theory-meets-bayesian-inference}, it may nevertheless be worthwhile to investigate whether these training objectives lead to better test MSE in examples. A drawback is that those assumptions are usually impossible to verify, thus the generalization guarantees are not comparable. Note that the design of a loss function is dependent on the application domain and there is no ubiquitous choice across all settings.  In many safety-critical applications, small deviations are tolerable whereas larger deviations are all equally catastrophic, thus a 0-1-loss as ours and a rigorous bound on it can be more useful than the MSE test error.


While in this work we focussed on regression tasks, the same strategy of optimizing a generalization bound can also be applied to learn GPs for binary and categorical outputs. Note that the true $KL$-term in this setting has so far been merely upper bounded by its regression proxy \cite{seeger-classification-paper}, and it would be interesting to develop better bounds on the classification complexity term. Lastly, it may be worthwhile to use other or more general sparse GPs within our PAC-Bayesian learning method, such as free-form \cite{scalable-variational-gaussian-process-classification} or even more general GPs \cite{more-general-than-free-form}.





\subsubsection*{Acknowledgments}
We would like to thank Duy Nguyen-Tuong, Martin Schiegg, and Michael Schober for helpful discussions and proofreading.


\newpage

\begin{center}
{\Large{Supplementary material for}}

{\LARGE{\textbf{Learning Gaussian Processes by}}}

{\LARGE{\textbf{Minimizing Generalization Bounds}}}
\end{center}

\appendix

\section{Inverse binary KL-divergence and its derivatives, Pinsker's inequality}\label{klinv-derivatives-appendix}
\begin{figure}[h!]
\begin{center}\includegraphics[trim={3.7cm 9.4cm 4cm 10cm},clip=true,scale=0.7]{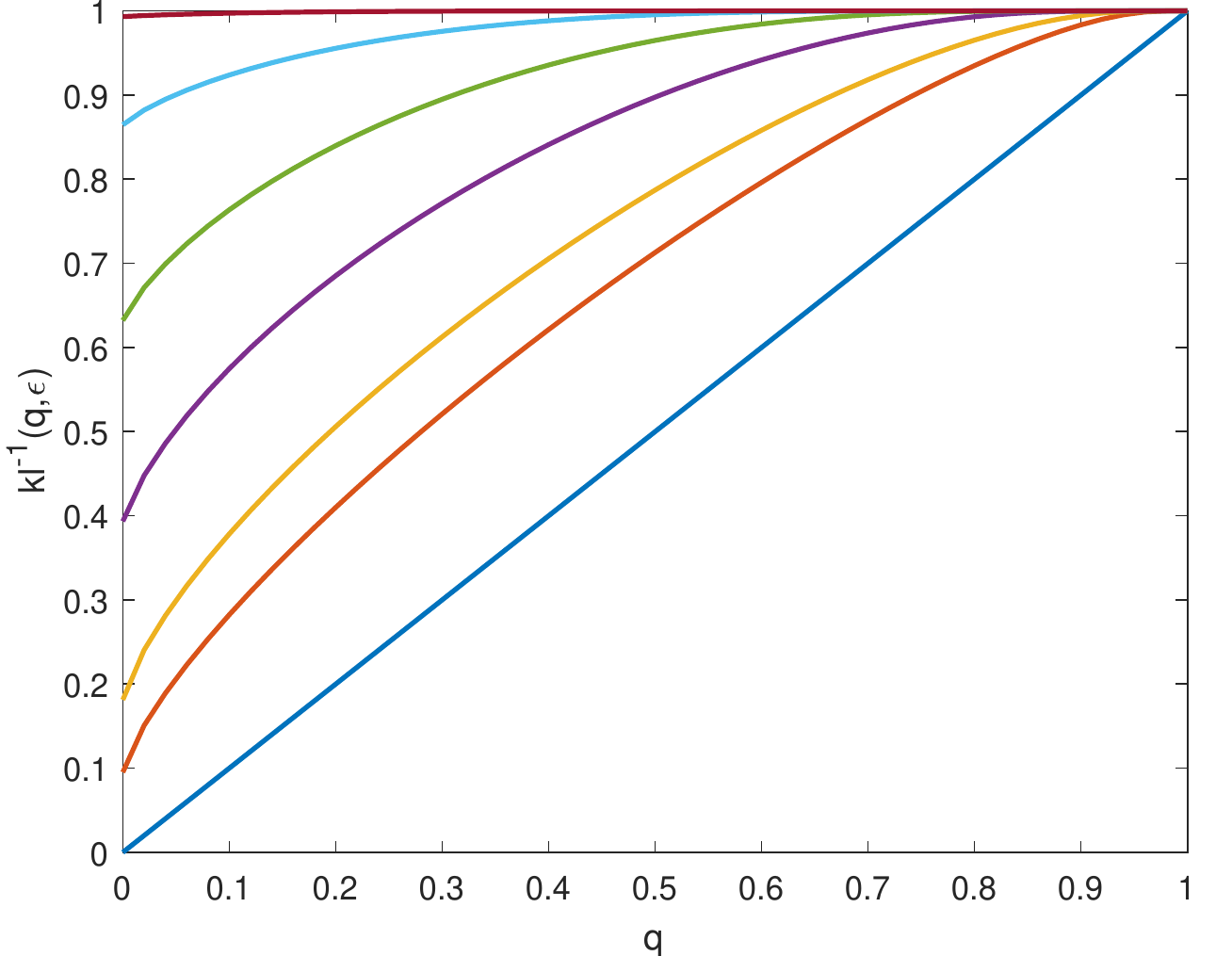}\end{center}  
\caption{\textbf{Inverse binary KL-divergence.} 
The figure shows plots of $\klinv(q,\varepsilon)$ for $\varepsilon\in\{0,0.1,0.2,0.5,1,2,5\}$ in different colors, the curves for larger $\varepsilon$ lying higher. For $\varepsilon=0$ it is $\klinv(q,\varepsilon=0)=q$ (staight blue line). At $q=0$ the curves start at $\klinv(q=0,\varepsilon)=1-e^{-\varepsilon}$. At $q=1$ we have $\klinv(q=1,\varepsilon)=1$ for any $\varepsilon\geq0$.   \label{klinv-plot}   }
\end{figure}

The function $\klinv(q,\varepsilon)\in[q,1]$, defined in Eq.\ (\ref{define-binary-kl}), can easily be computed numerically for any $q\in[0,1)$, $\varepsilon\in[0,\infty)$ to any desired accuracy $\Delta>0$ via the bisection method, since the function $\kl{q}{p}=q\ln\frac{q}{p}+(1-q)\ln\frac{1-q}{1-p}$ is strictly monotonically increasing in $p\in[q,1]$ from $0$ to $\infty$ (for $q=1$ or $\varepsilon=\infty$, we set $\klinv(q,\varepsilon):=1$). Note that the monotonicity in $p$ implies further that $\klinv(q,\varepsilon)$ is monotonically increasing in $\varepsilon\in[0,\infty]$. Fig. \ref{klinv-plot} shows a plot of $\klinv(q,\varepsilon)$ for various values of $\epsilon\geq0$, and states a few special function values of $\klinv(q,\varepsilon)$. By Pinsker's inequality, it holds that $\kl{q}{p}\geq2|p-q|^2$, which implies that $\klinv(q,\varepsilon)\leq q+\sqrt{\varepsilon/2}$.

When applying gradient descent on the RHS of the generalization bound (\ref{pac-bayes-union-bound-eq}) (which includes (\ref{pac-bayes-bound-eq}) as a special case), as we propose and do, one further needs, besides the evaluation of $\klinv$, also the derivatives of $\klinv$ w.r.t.\ both of its arguments. These can be easily derived by differentiating the identity $\kl{q}{\klinv(q,\varepsilon)}=\varepsilon$ w.r.t.\ $q$ and $\varepsilon$, plugging in the easily computed derivatives of $\kl{q}{p}=q\ln\frac{q}{p}+(1-q)\ln\frac{1-q}{1-p}$. The result is:
\begin{align}
\frac{\partial\,\klinv(q,\varepsilon)}{\partial q}&=\frac{\ln\frac{1-q}{1-\klinv(q,\varepsilon)} - \ln\frac{q}{\klinv(q,\varepsilon)}}{\frac{1-q}{1-\klinv(q,\varepsilon)} - \frac{q}{\klinv(q,\varepsilon)}},\label{partial-kl-1-1}\\
\frac{\partial\,\klinv(q,\varepsilon)}{\partial \varepsilon}&=\frac{1}{\frac{1-q}{1-\klinv(q,\varepsilon)} - \frac{q}{\klinv(q,\varepsilon)}}.\label{partial-kl-1-2}
\end{align}
The derivative of the RHS of (\ref{pac-bayes-union-bound-eq}) with respect to parameters $\xi$ (which may include the hyperparameters $\theta$, the noise level $\sigma_n^2$, the inducing points $z_i$, or any other parameters of $P$ and $Q$ such as $a_M,B_{MM}$, or $\alpha$ from Section \ref{sparse-GP-subsection}) thus reads:
\begin{equation}\label{deriv-kl-1-chain}
\begin{split}
\frac{d}{d\xi}\,&\klinv\left(R_S(Q_\xi),\frac{\KL{Q_\xi}{P_\xi}+\ln\frac{1}{p_\xi}+\ln\frac{2\sqrt{N}}{\delta}}{N}\right)=\\
&=\left(\left.\frac{\partial\,\klinv(q,\varepsilon)}{\partial q}\right|_{\begin{smallmatrix}q=R_S(Q_\xi)\\ \varepsilon=(\KL{Q_\xi}{P_\xi}+\ln\frac{2\sqrt{N}}{\delta p_\xi})/N\end{smallmatrix}}\right)\,\cdot\,\frac{d}{d\xi}R_S(Q_\xi)\\
&\ \ \ \ +\left(\left.\frac{\partial\,\klinv(q,\varepsilon)}{\partial \varepsilon}\right|_{\begin{smallmatrix}q=R_S(Q_\xi)\\ \varepsilon=(\KL{Q_\xi}{P_\xi}+\ln\frac{2\sqrt{N}}{\delta p_\xi})/N\end{smallmatrix}}\right)\,\cdot\,\frac{d}{d\xi}\frac{\KL{Q_\xi}{P_\xi}+\ln\frac{1}{p_\xi}+\ln\frac{2\sqrt{N}}{\delta}}{N},
\end{split}
\end{equation}
using the partial derivatives of $\klinv$ in parentheses from (\ref{partial-kl-1-1})--(\ref{partial-kl-1-2}).

We use the expression (\ref{deriv-kl-1-chain}) for our gradient-based optimization of the parameters $\xi$ in Sect.\ \ref{experiments-section}. Note that we treat all components of $\xi$ as continuous parameters during this optimization, despite the fact that the hyperparameters $\theta\in\Theta$ for the prior $P_\theta$ have to come from a \emph{countable} set $\Theta$ (see around Eq.\ (\ref{pac-bayes-union-bound-eq}) and App.\ \ref{KL-theta-appendix}). It is only after the optimization that we discretize all of those parameters $\theta$ onto a pre-defined grid (chosen before seeing the training sample $S$), as described in Sect.\ \ref{full-GP-section}.

Note that $\frac{d}{d\xi}\KL{Q_\xi}{P_\xi}$ in (\ref{deriv-kl-1-chain}) can be computed analytically in our proposed methods PAC-GP and PAC-SGP by using standard matrix algebra (e.g., \cite[App.\ A]{rasmussen-williams-book}) for differentiating the matrix-analytic expressions of $\KL{Q_\xi}{P_\xi}$ in (\ref{make-KL-finite-dim}) or (\ref{KL-for-sparse-GP}) (possibly with the parametrization (\ref{BMM-sparse-eqn})). Furthermore, the derivative $\frac{d}{d\xi}\ln\frac{1}{p_\xi}=-\frac{1}{p_\xi}\frac{d}{d\xi}p_\xi$ is easily computed for common distributions $p_\theta$ (Sect.\ \ref{PAC-Bayesian-thm-section}), again treating $\xi$ first as a continuous parameter in the optimization as explained in the previous paragraph; in our paper, we always discretize the hyperparameter set $\Theta$ to be finite and choose $p_\xi=\frac{1}{|\Theta|}$ as the uniform distribution, so $p_\xi$ is independent of $\xi$ and $\frac{d}{d\xi}\ln\frac{1}{p_\xi}=0$. Lastly, we show in App.\ \ref{derivatives-Phi-appendix} how to effectively compute $\frac{d}{d\xi}R_S(Q_\xi)$ in the expression (\ref{deriv-kl-1-chain}) for relevant loss functions $\ell$.

To our knowledge, the parameter-free PAC-Bayes bound from Theorem \ref{pac-bayes-theorem} or Eq.\ (\ref{pac-bayes-union-bound-eq}) has never before been used for learning, as we do in our paper here, ostensibly due to the perceived difficulty of handling the derivatives of $\klinv$ \cite{nonvacuous-generalization-bounds-snn}. Instead, when a PAC-Bayes bound was used to guide learning in prior works \cite{pac-bayes-learning-of-linear-classifiers,nonvacuous-generalization-bounds-snn}, then a simple sum of $R_S(Q)$ and a penalty term involving $\KL{Q}{P}$ and $\log\frac{1}{p_\theta}$ was employed as an upper bound, either obtained from alternative PAC-Bayes theorems \cite{catoni-thermodyn-of-ml,pac-bayes-learning-of-linear-classifiers} or from loosening the upper bound in Eq.\ (\ref{pac-bayes-union-bound-eq}) to an expression of the form $R_S(Q)+\sqrt{\big(\KL{Q}{P}+\ln\frac{2\sqrt{N}}{\delta}\big)/(2N)}$ by a use of Pinsker's inequality \cite{nonvacuous-generalization-bounds-snn} or by looser derivations (some of which are mentioned in \cite{pac-bayes-learning-of-linear-classifiers,pac-bayesian-bounds-based-on-the-renyi-divergence}). We show in our work how to perform the learning directly with the $\klinv$-bound (\ref{pac-bayes-union-bound-eq}) using the derivative from Eq.\ (\ref{deriv-kl-1-chain}), and demonstrate that its optimization is robust and stable and has better performance than the optimization of looser bounds (see Sect.\ \ref{experiments-section}).

\newpage

\section{Proof of Eq.\ (\ref{pac-bayes-union-bound-eq}) --- KL-divergence inequality and union bound}\label{KL-theta-appendix}
Let $\Theta$ be a \emph{countable} set (i.e.\ a finite set or a countably infinite set), and let $p_\theta$ be any probability distribution over its elements $\theta\in\Theta$. Further, let $P^\theta$ be a family of probability distributions indexed by the $\theta\in\Theta$, and define their mixture $P:=\sum_{\theta'} p_{\theta'} P^{\theta'}$. Then it holds for each $\theta\in\Theta$ and $Q$:
\begin{align}
\KL{Q}{P}&=\int dx\,Q(x)\ln\frac{Q(x)}{P(x)}\label{KLQP-appendix-eqn}\\
&=\int dx\,Q(x)\ln\frac{Q(x)}{\sum_{\theta'}p_{\theta'}P^{\theta'}(x)}\nonumber\\
&\leq\int dx\,Q(x)\ln\frac{Q(x)}{p_\theta P^\theta(x)}\label{simple-inequality-in-KL-derivation}\\
&=\int dx\,Q(x)\ln\frac{1}{p_\theta}\,+\,\int dx\,Q(x)\ln\frac{Q(x)}{P^\theta(x)}\nonumber\\
&=\ln\frac{1}{p_\theta}\,+\,\KL{Q}{P^\theta}.\label{KLQPtheta-appendix-eqn}
\end{align}
The inequality (\ref{simple-inequality-in-KL-derivation}) follows from the simple fact that the sum $\sum_{\theta'}p_{\theta'}P^{\theta'}(x)$ contains only nonnegative terms and is therefore at least as large as any of its summands, $\sum_{\theta'}p_{\theta'}P^{\theta'}(x)\geq p_\theta P^\theta(x)$, together with the monotonicity of the logarithm $\ln$. This inequality would not generally hold when $\sum_{\theta'}$ were replaced by an integral $\int_{\theta'}$ over a continuous index $\theta'$, which explains the requirement of a \emph{countable} index set $\Theta$. The inequality $\KL{Q}{P}\leq\ln\frac{1}{p_\theta}+\KL{Q}{P^\theta}$ holds also for $p_\theta=0$ with the interpretation $\ln\frac{1}{0}=\infty$.

The inequality $\KL{Q}{P}\leq\ln\frac{1}{p_\theta}+\KL{Q}{P^\theta}$ from (\ref{KLQP-appendix-eqn})--(\ref{KLQPtheta-appendix-eqn}), together with fact that $\klinv$ is monotonically increasing in its second argument, shows how to obtain Eq.\ (\ref{pac-bayes-union-bound-eq}) from Theorem \ref{pac-bayes-theorem}.

\begin{remark}
Using the value $\KL{Q}{P}$ with $P=\sum_{\theta'}p_{\theta'}P^{\theta'}$ directly in (\ref{pac-bayes-bound-eq}) would of course yield a better bound than (\ref{pac-bayes-union-bound-eq}), but this $\KL{Q}{P}$ is generally difficult to evaluate, e.g.\ when $P$ is a mixture of Gaussians. Furthermore, the alternative bound $\KL{Q}{\sum_{\theta'} p_{\theta'} P^{\theta'}}\leq\sum_{\theta'} p_{\theta'} \KL{Q}{P^{\theta'}}$, coming from convexity of $KL$, would require the value of $\KL{Q}{P^{\theta'}}$ for each $\theta'\in\Theta$; but $\KL{Q}{P^{\theta'}}$ cannot be computed by the automatic method of Sections \ref{full-GP-section}--\ref{sparse-GP-subsection} when $Q$ originates from different hyperparameters than $P^{\theta'}$.
\end{remark}

As an alternative derivation of Eq.\ (\ref{pac-bayes-union-bound-eq}) from Theorem \ref{pac-bayes-theorem}, one may use the ordinary union bound argument: For a given probability distribution $p_\theta$ on the countable set $\Theta$ and a given $\delta\in(0,1]$, define $\delta_\theta:=\delta p_\theta$. Now consider the statement of Theorem \ref{pac-bayes-theorem} for each prior $P^\theta$ individually with confidence parameter $\delta_\theta$; this gives that, for each $\theta\in\Theta$, the statement
\begin{align}
\forall Q:\quad R(Q)&\leq \klinv\left(R_S(Q),\frac{\KL{Q}{P^\theta}+\ln\frac{2\sqrt{N}}{\delta_\theta}}{N}\right)\nonumber\\
&=\klinv\left(R_S(Q),\frac{\KL{Q}{P^\theta}+\ln\frac{1}{p_\theta}+\ln\frac{2\sqrt{N}}{\delta}}{N}\right)\nonumber
\end{align}
fails with probability at most $\delta_\theta$ (over $S\sim\mu^N$). By the union bound, the statement fails for one $\theta\in\Theta$ with probability at most $\sum_\theta \delta_\theta=\sum_\theta \delta p_\theta=\delta\sum_\theta p_\theta=\delta\cdot1=\delta$. Thus, the statement of Eq.\ (\ref{pac-bayes-union-bound-eq}) (containing the quantifier $\forall\theta$) holds with probability at least $1-\delta$ over $S\sim\mu^N$.

\newpage

\section[Loss functions, the empirical risk R\_S(Q), and its gradient]{Loss functions, the empirical risk $R_S(Q)$, and its gradient}\label{derivatives-Phi-appendix}
\begin{figure}
\begin{center}\includegraphics[trim={3.7cm 13cm 4cm 13.7cm},clip=true,scale=0.9]{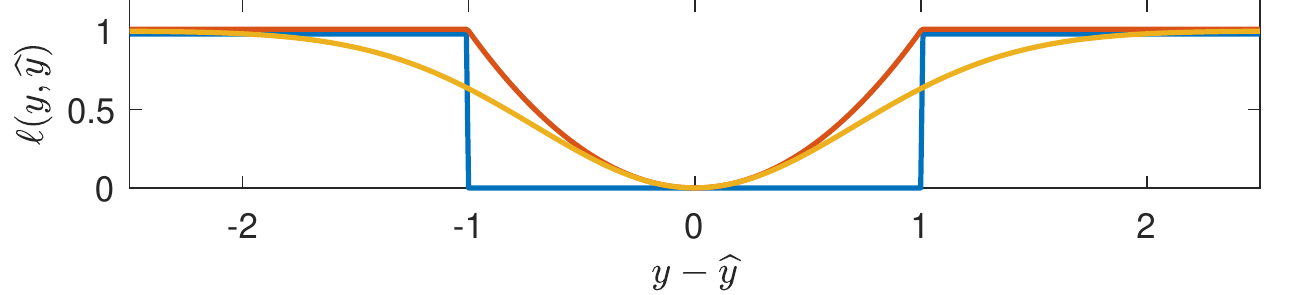}\end{center}  
\caption{\textbf{Various regression loss functions.} Shown are three bounded loss functions $\ell$, which are appropriate for the regression setting and which allow for an effective computation of the empirical risk $R_S(Q)$ when $Q$ is a GP. Each of these three functions $\ell(y,\widehat{y})$ depends only on the absolute deviation $y-\widehat{y}$ (horizontal axis), and contains a scale parameter $\varepsilon>0$ which is set to $\varepsilon=1$ in the plots: $\ell_\ii(y,\widehat{y})=\ii_{|y-\widehat{y}|>\varepsilon}=\ii_{\widehat{y}\notin[y-\varepsilon,y+\varepsilon]}$ (blue), which we use in our experiments, $\ell_2(y,\widehat y)=\min\{[(y-\widehat y)/\varepsilon]^2,1\}$ (red), and $\ell_{\exp}(y,\widehat y)=1-\exp[-((y-\widehat y)/\varepsilon)^2]$ (yellow). \label{loss-functions-plot}  }
\end{figure}

Our proposed method requires the empirical risk $R_S(Q)$ on the training set $S=\{(x_i,y_i)\}_{i=1}^N$ (see Sect.\ \ref{setup-section}) to be computed effectively for any considered distribution $Q$, along with its gradient $\frac{d}{d\xi}R_S(Q_\xi)$ for gradient-based optimization. We show here that this can be done for many interesting loss functions $\ell$ when $Q$ is a Gaussian Process, including the following (see Fig.\ \ref{loss-functions-plot} for illustration):
\begin{align}
\ell_\ii(y,\widehat{y})&=\ii_{|y-\widehat{y}|>\varepsilon}=\ii_{\widehat{y}\notin[y-\varepsilon,y+\varepsilon]}\,,\label{ii-loss-function}\\
\ell_2(y,\widehat y)&=\min\{((y-\widehat y)/\varepsilon)^2,1\}\,,\label{loss-function-2}\\
\ell_{\exp}(y,\widehat y)&=1-\exp[-((y-\widehat y)/\varepsilon)^2]\,,\label{loss-function-exp}\\
\ell_\pm(y,\widehat{y})&=\ii_{\widehat y\notin[r_-(y),r_+(y)]}\,,\label{loss-function-pm}
\end{align}
where $\varepsilon>0$ is a scale parameter to be chosen for the first three, and $r_\pm(y)$ are functions to be specified for the last. Note that $\ell_\ii$ specifies an \emph{additive} accuracy goal $\pm\varepsilon$ and was used in our experiments (Sect.\ \ref{experiments-section}), whereas we have suggested $\ell_2$ and $\ell_{\exp}$ as more deviation-sensitive (yet bounded) loss functions that may yield better results on the MSE error (see Sect.\ \ref{conclusion-section}, and Sect.\ \ref{experiments-section}). The loss function $\ell_\pm$ generalizes $\ell_\ii$ (which is obtained by using the functions $r_\pm(y):=y\pm\varepsilon$, see Sect.\ \ref{setup-section}), but could also be used to specify \emph{relative} accuracy goals, e.g.\ setting $r_\pm(y):=y\pm\varepsilon|y|$. More deviation-sensitive relative loss functions are possible as well, e.g.\ $\ell(y,\widehat{y}):=\max\left\{\left|\frac{\widehat{y}-y}{y}\right|,1\right\}$, which we do not treat here but which allows similarly effective computation as the other ones.

Let us denote by $\widehat{m}(x)$ and $\widehat{\sigma}^2(x)$ the predictive mean and variance of the predictive GP $Q$. In our work we use the two forms (\ref{usual-GPR-posterior}) (PAC-GP) and (\ref{posterior-general-sparse-GP}) (sparse PAC-SGP); in the latter case we e.g.\ have:
\begin{align}
\widehat{m}(x)&=m(x)+k_M(x)K_{MM}^{-1}(a_M-m_M),\\
\widehat{\sigma}^2(x)&=K(x,x')-k_M(x)K_{MM}^{-1}[K_{MM}-B_{MM}]K_{MM}^{-1}k_M(x')^T.
\end{align}
We denote by $\widehat{m}_i:=\widehat{m}(x_i)$, $\widehat{\sigma}^2_i:=\widehat{\sigma}^2(x_i)$ the predictive mean and variance at the training inputs. The empirical risk $R_S(Q)$ from (\ref{stoch-empirical-risk}) then reduces to a sum of one-dimensional integrals containing a Gaussian density:
\begin{align}
R_S(Q)&=\frac{1}{N}\sum_{i=1}^N\mathbb E_{h\sim Q}\big[\ell\big(y_i,h(x_i)\big)\big]=\frac{1}{N}\sum_{i=1}^N\mathbb E_{v\sim Q(x_i)}\big[\ell\big(y_i,v\big)\big]\\
&=\frac{1}{N}\sum_{i=1}^N\int dv\,{\mathcal N}(v \mid \widehat{m}_i,\widehat{\sigma}_i^2)\,\ell(y_i,v).\label{compute-RS-in-appendix}
\end{align}
The last integral can be evaluated for each of the loss functions (\ref{ii-loss-function})--(\ref{loss-function-pm}):
\begin{align}
\int dv\,{\mathcal N}(v \mid \widehat{m}_i,\widehat{\sigma}_i^2)\,\ell_\ii(y_i,v)&=\Phi\left(\frac{y_i-\varepsilon-\widehat{m}_i}{\widehat{\sigma}_i}\right)+1-\Phi\left(\frac{y_i+\varepsilon-\widehat{m}_i}{\widehat{\sigma}_i}\right)\,,\label{integral-1-app}\\
\int dv\,{\mathcal N}(v \mid \widehat{m}_i,\widehat{\sigma}_i^2)\,\ell_2(y_i,v)&=\left(1-\frac{(y_i-\widehat{m}_i)^2+\widehat{\sigma}_i^2}{\varepsilon^2}\right)\left(\Phi\left(\frac{y_i-\varepsilon-\widehat{m}_i}{\widehat{\sigma}_i}\right)-\Phi\left(\frac{y_i+\varepsilon-\widehat{m}_i}{\widehat{\sigma}_i}\right)\right)\nonumber\\
&\quad+1-\frac{\widehat{\sigma}_i}{\sqrt{2\pi}\varepsilon^2}(y_i-\varepsilon-\widehat{m}_i)e^{-(y_i+\varepsilon-\widehat{m}_i)^2/(2\widehat{\sigma}_i^2)}\\
&\quad\ \ \ \ \ \ \ -\frac{\widehat{\sigma}_i}{\sqrt{2\pi}\varepsilon^2}(y_i+\varepsilon-\widehat{m}_i)e^{-(y_i-\varepsilon-\widehat{m}_i)^2/(2\widehat{\sigma}_i^2)},\nonumber\\
\int dv\,{\mathcal N}(v \mid \widehat{m}_i,\widehat{\sigma}_i^2)\,\ell_{\exp}(y_i,v)&=1-\frac{1}{\sqrt{1+\frac{2\widehat{\sigma}_i^2}{\varepsilon^2}}}\exp\left[-\frac{(y_i-\widehat{m}_i)^2}{2\widehat{\sigma}_i+\varepsilon^2}\right]\,,\\
\int dv\,{\mathcal N}(v \mid \widehat{m}_i,\widehat{\sigma}_i^2)\,\ell_\pm(y_i,v)&=\Phi\left(\frac{r_-(y_i)-\widehat{m}_i}{\widehat{\sigma}_i}\right)+1-\Phi\left(\frac{r_+(y_i)-\widehat{m}_i}{\widehat{\sigma}_i}\right)\,,\label{integral-4-app}
\end{align}
where by
\begin{align}
\Phi(z):=\int_{-\infty}^{z}\frac{1}{\sqrt{2\pi}}e^{-t^2/2}dt
\end{align}
we denote the cumulative distribution function of a standard normal, which is implemented in most computational packages. Plugging the expressions (\ref{integral-1-app})--(\ref{integral-4-app}) into (\ref{compute-RS-in-appendix}) shows how $R_S(Q)$ can be computed.

With the above expressions one can also compute gradients of $R_S(Q)=R_S(Q_\xi)$ effectively for gradient-based optimization: When $Q=Q_\xi$ depends on parameters $\xi$ (such as hyperparameters $\theta$, noise $\sigma_n$, inducing points $\{z_i\}$, or any other free-form parameters $a_M$, $B_{MM}$ or $\alpha$ from Sect.\ \ref{sparse-GP-subsection}), then $\widehat{m}(x)=\widehat{m}^\xi(x)$ and $\widehat{\sigma}(x)=\widehat{\sigma}^\xi(x)$ depend on $\xi$ as well through explicit expressions, via (\ref{usual-GPR-posterior}) and (\ref{posterior-general-sparse-GP}). One can thus compute the gradients $\frac{d}{d\xi}\widehat{m}^\xi_i=\frac{d}{d\xi}\widehat{m}^\xi(x_i)$ and $\frac{d}{d\xi}\widehat{\sigma}^\xi_i=\frac{d}{d\xi}\widehat{\sigma}^\xi(x_i)$ analytically, using standard matrix analysis (e.g.\ \cite[App.\ A]{rasmussen-williams-book}). With these gradients and the above expressions (\ref{integral-1-app})--(\ref{integral-4-app}) it is easy to compute $\frac{d}{d\xi}R_S(Q_\xi)$ for the above loss function; e.g.\ for $\ell_\ii$ from (\ref{ii-loss-function}) used in our experiments:
\begin{align}
\frac{d}{d\xi}R_S^\ii(Q_\xi)=\frac{1}{N}\sum_{i=1}^N \left[\left(\frac{d}{d\xi}\frac{y_i-\varepsilon-\widehat{m}_i}{\widehat{\sigma}_i}\right)e^{-\frac{1}{2}\left(\frac{y_i-\varepsilon-\widehat{m}_i}{\widehat{\sigma}_i}\right)^2} - \left(\frac{d}{d\xi}\frac{y_i+\varepsilon-\widehat{m}_i}{\widehat{\sigma}_i}\right)e^{-\frac{1}{2}\left(\frac{y_i+\varepsilon-\widehat{m}_i}{\widehat{\sigma}_i}\right)^2}\right]
\label{example-of-RS-gradient}
\end{align}
where we used that $\frac{d}{dz}\Phi(z)=\frac{1}{\sqrt{2\pi}}e^{-z^2/2}$.

Lastly, for the purpose of gradient-based optimization of the objective from Theorem \ref{pac-bayes-theorem} or Eq.\ (\ref{pac-bayes-union-bound-eq}), one does not really need to compute the exact $R_S(Q)$ as a sum over $N$ training examples, which is possibly a large number. Rather, one could do mini-batches of size $B\ll N$ and obtain a stochastic estimate
\begin{align}
R_S(Q)\,\approx\,\frac{1}{B}\sum_{i=1}^B\int dv\,{\mathcal N}(v \mid \widehat{m}_i,\widehat{\sigma}_i^2)\,\ell(y_i,v)=:\widehat{R}_B(Q),
\end{align}
where the sum runs over one mini-batch selected from the $N$ training points randomly or in cyclic order. (Hoeffding's inequality gives that $|R_S(Q)-\widehat{R}_B(Q)|\lesssim\sqrt{\frac{1}{2B}\ln\frac{2}{\delta'}}$ holds with probability $\geq1-\delta'$ over mini-batches. While this statement could be incorporated into a version of Theorem \ref{pac-bayes-theorem} or Eq.\ (\ref{pac-bayes-union-bound-eq}) that is expressed in terms of $\widehat{R}_B(Q)$ instead of $R_S(Q)$, we propose stochastic estimates $\widehat{R}_B(Q)$ only during the optimization procedure and suggest a full computation of $R_S(Q)$ for the final evaluation of the generalization bound.) Similarly, the exact gradient $\frac{d}{d\xi}R_S(Q_\xi)$ is a sum over $N$ training examples (e.g., (\ref{example-of-RS-gradient})), so one can approximate it in the same way by mini-batches to obtain a faster stochastic estimate of the gradient which is often sufficient for optimization.

\newpage

\section{Training objectives of other GP methods}\label{other-objective-functions-section}
Here we contrast our proposed learning objective (\ref{pac-bayes-union-bound-eq}) with those of other common GP methods, to which we compare in the experiments (Sect.\ \ref{experiments-section}).

In standard full GP regression learning \cite{rasmussen-williams-book} one selects those prior hyperparameters $\theta$ and noise level $\sigma_n$ which maximize the data likelihood $p(\trainoutputs \mid \theta,\sigma_n)=\mathcal N(\trainoutputs \mid m_N,K_{NN}+\sigma_n^2\ii)$
under the \emph{prior} GP. This corresponds to the minimization objective
\begin{equation}\label{full-gp-likelihood}
-\ln p(\trainoutputs \mid \theta,\sigma_n)=\frac{1}{2}\ln\det[K_{NN}+\sigma_n^2\ii]+\frac{N}{2}\ln(2\pi)+\frac{1}{2}(\trainoutputs-m_N)^T(K_{NN}+\sigma_n^2\ii)^{-1}(\trainoutputs-m_N).
\end{equation}
The optimal $\theta$, $\sigma_n$ are then used in (\ref{usual-GPR-posterior}) to make predictions.

The sparse-GP methods FITC \cite{snelson-spgp}, VFE \cite{titsias09}, and DTC \cite{seeger-dtc} adjust $\theta$, $\sigma_n$, and the $M$ inducing inputs $\{z_i\}$ by minimizing the objective \cite{bauer-understanding-sparse-GP-approximations}
\begin{equation}
\begin{split}
\mathcal F&=\frac{1}{2}\ln\det\big[K_{NM}K_{MM}^{-1}K_{MN}+\sigma_n^2\ii+G\big]+\frac{N}{2}\ln(2\pi)+\frac{1}{2\sigma_n^2}{\rm tr}[T]\\
&\ \ \ \ +\frac{1}{2}(\trainoutputs-m_N)^T\big(K_{NM}K_{MM}^{-1}K_{MN}+\sigma_n^2\ii+G^\theta\big)^{-1}(\trainoutputs-m_N),
\end{split}
\end{equation}
where $G_{\rm FITC}=T_{VFE}={\rm diag}(K_{NN}-K_{NM}K_{MM}^{-1}K_{MN})$ and $G_{\rm VFE}=G_{\rm DTC}=T_{FITC}=T_{DTC}=0$. For DTC and FITC, $\mathcal F$ are the negative log likelihoods of approximate prior models \cite{unifying-view-sparse-gp-approx,snelson-spgp}, whereas $\mathcal F$ equals the exact negative log likelihood plus the KL-divergence $\KL{Q}{\widetilde{Q}}$ between $Q$ and the exact Bayesian posterior $\widetilde{Q}$ obtained from the Bayesian prior $P$. VFE and DTC make predictions $Q$ by using (\ref{BMM-sparse-eqn}) with $\alpha=0$ in (\ref{posterior-general-sparse-GP}), whereas FITC sets $\alpha=1$.

One can compare the above expressions to the $\KL{Q}{P}$ term in the PAC-Bayes bound (\ref{pac-bayes-union-bound-eq}). For our full-GP training, $\KL{Q}{P}$ is given in (\ref{non-sparse-KL-formula}):
\begin{align}
\KL{Q}{P}&=-\frac{1}{2}\ln\det\big[K_{NN}^{-1}(K_{NN}-K_{NN}(K_{NN}+\sigma_n^2\ii)^{-1}K_{NN})\big]\nonumber\\
&\ \ \ \ \ +\frac{1}{2}{\rm tr}[K_{NN}^{-1}(K_{NN}-K_{NN}(K_{NN}+\sigma_n^2\ii)^{-1}K_{NN})\big]\,-\,\frac{N}{2}\nonumber\\
&\ \ \ \ \ +\frac{1}{2}(\trainoutputs-m_N)^T(K_{NN}+\sigma_n^2\ii)^{-1}K_{NN}(K_{NN}+\sigma_n^2\ii)^{-1}(\trainoutputs-m_N),\nonumber\\
&=\frac{1}{2}\ln\det\big[K_{NN}+\sigma_n^2\ii\big]-\frac{N}{2}\ln\sigma_n^2-\frac{1}{2}{\rm tr}\big[K_{NN}(K_{NN}+\sigma_n^2\ii)^{-1}\big]\nonumber\\
&\ \ \ \ \ +\frac{1}{2}(\trainoutputs-m_N)^T(K_{NN}+\sigma_n^2\ii)^{-1}K_{NN}(K_{NN}+\sigma_n^2\ii)^{-1}(\trainoutputs-m_N),\nonumber\\
&=\frac{1}{2}\sum_{i=1}^N\Big[\ln\frac{\lambda_i+\sigma_n^2}{\sigma_n^2}-\frac{\lambda_i}{\lambda_i+\sigma_n^2}\Big]~+~\frac{1}{2}\sum_{i=1}^N \frac{\lambda_i}{(\lambda_i+\sigma_n^2)^2}(e_i\cdot(y-m_N))^2,\nonumber
\end{align}
where $\lambda_i\in\mathbb R$ are the eigenvalues of $K_{NN}$ and $e_i\in\mathbb R^N$ corresponding orthonormal eigenvectors. For our sparse-GP training with a ``free-form'' sparsification $Q(f_M)=\mathcal N(f_M \mid a_M,B_{MM})$ with free $a_M$, $B_{MM}$, it is from (\ref{KL-for-sparse-GP}):
\begin{align}
\KL{Q}{P}=\KL{Q(f_M)}{P(f_M)}=&-\frac{1}{2}\ln\det\big[B_{MM}K_{MM}^{-1}\big]+\frac{1}{2}{\rm tr}\big[B_{MM}K_{MM}^{-1}\big]-\frac{M}{2}  \nonumber\\
&+\frac{1}{2}(a_M-m_M)^T K_{MM}^{-1} (a_M-m_M),\nonumber
\end{align}
which via $a_M= K_{MM}Q_{MM}^{-1}K_{MN}(\alpha\Lambda+\sigma_n^2\ii)^{-1}\trainoutputs$, $B_{MM}=K_{MM}Q_{MM}^{-1}K_{MM}$ from (\ref{BMM-sparse-eqn}) with $\alpha=1$ can be particularized for the FITC parametrization used in our PAC-SGP work.


\newpage

\section{Experiment: predictive distributions of sparse GPs, and overfitting}
\label{sec:exper-comp-sparse}

\begin{figure}[h!]
  \centering
\includegraphics[width=\textwidth]{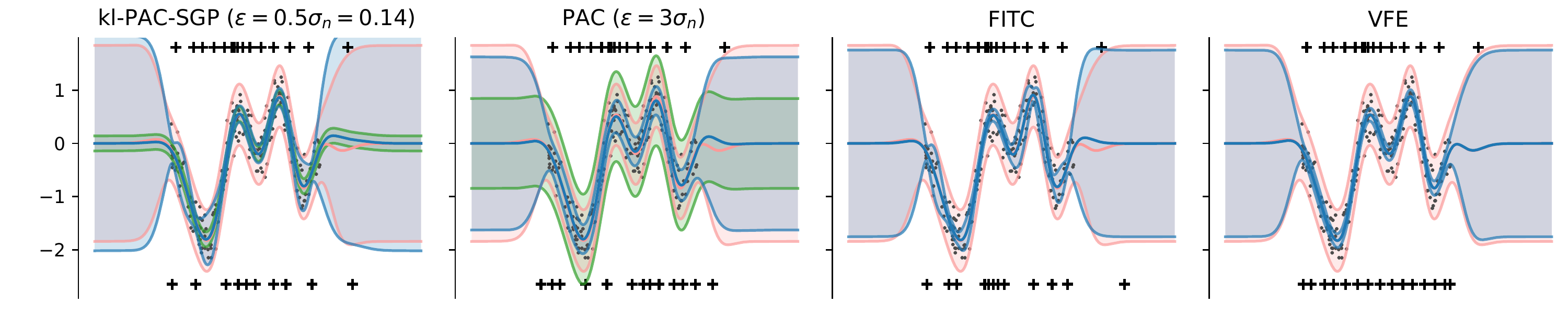}
  \caption{\textbf{Comparison of predictive distributions.} In each plot, we show the predictive distribution from a full-GP in red (mean $\pm$ twice the predictive variance), fitted to the data (black dots). The blue distributions (mean $\pm$ twice the predictive variance) in the first two plots are obtained form our sparse kl-PAC-SGP with two different values of $\varepsilon$ (chosen relative to the noise level $\sigma_n=0.28$ of the full-GP), the third shows the predictive distribution from FITC, the fourth from VFE. For the PAC-GP variants, we additionally plotted the $\varepsilon$-band as in Fig.\ \ref{snelson-fig}. As in Fig. \ref{snelson-fig}, the crosses show the inducing point positions before and after training. \label{fig:snelson-ftic-vfe-comp} }
\end{figure}

To compare the predictive distributions of common sparse GPs to the predictive distribution obtained from our sparse PAC-SGP method (Sect.\ \ref{sparse-GP-subsection}) optimized with the PAC-Bayesian bound (\ref{pac-bayes-union-bound-eq}), we trained FITC \cite{snelson-spgp} and VFE \cite{titsias09} on the same dataset used in Fig.\ \ref{snelson-fig}, which was also used in \cite{snelson-spgp,titsias09} for a comparison of methods.  It can be seen in Fig.\ \ref{fig:snelson-ftic-vfe-comp} that  especially for small $\varepsilon$ our PAC-SGP has a predictive distribution more similar to FITC, whereas for larger $\varepsilon$, the predictive distribution becomes closer to the full-GP, however not as close as VFE. Note that, for the full-GP, for FITC and for VFE we include the additive observation noise $\sigma_n^2$ in the predictive uncertainty in Fig.\ \ref{fig:snelson-ftic-vfe-comp}, whereas for our PAC-SGP variant we do not include additive observation noise $\sigma_n^2$, since this is not part of the predictive variance (see Eqs.\ (\ref{posterior-general-sparse-GP},\ref{BMM-sparse-eqn}), and similarly Eq.\ (\ref{usual-GPR-posterior}) for the non-sparse case); we instead plot the $\varepsilon$-band from the 0-1-loss function (green) around the predictive PAC-SGP mean. We further refer to the discussions in \cite{snelson-spgp,titsias09} concerning the same dataset.



\bigskip
\bigskip

\begin{figure}[h!]
  \centering
\includegraphics[width=\textwidth]{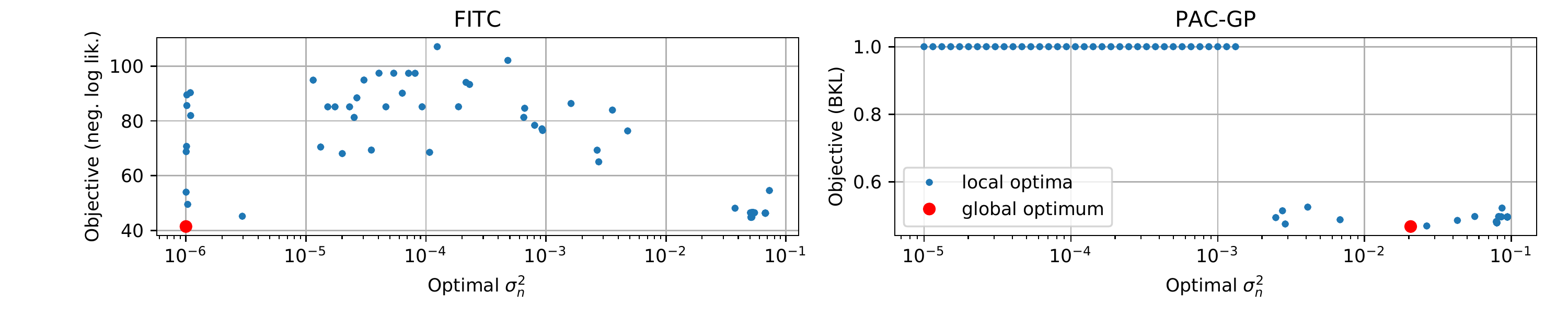}
  \caption{\textbf{Local minima of the optimization for different initializations.} Shown are the learned $\sigma_n^2$ and the achieved (local) minima for 100 different initializations of $\sigma_n^2\in[10^{-5},10^{+1}]$ for the FITC and kl-PAC-SGP methods trained on 100 out of the 200 datapoints of the 1D-dataset from \cite{snelson-spgp,titsias09}. See also \cite[Section 3.1]{bauer-understanding-sparse-GP-approximations}.  \label{fig:objective-scatter} }
\end{figure}

As a further comparison of our method with FITC, we now illustrate the well-known overfitting of the FITC method on pathological datasets \cite{bauer-understanding-sparse-GP-approximations} and show how our PAC-SGP method avoids it. The dataset for this demonstration consists of half of the datapoints (using every second one) of the above 1D-dataset \cite{snelson-spgp,titsias09}, similar to what was done in the comparison study in \cite[Section 3.1]{bauer-understanding-sparse-GP-approximations}. For 100 different initializations of $\sigma_n^2\in[10^{-5},10^{+1}]$ and the $M=8$ inducing inputs, we trained a FITC model and a kl-PAC-SGP model, minimizing the (approx.) negative log-likelihood for FITC and minimizing the BKL bound from Eq.\ (\ref{pac-bayes-union-bound-eq}) for kl-PAC-SGP (using the 0-1-loss function with $\varepsilon=0.6$, cf.\ Sect.\ \ref{experiments-section}). Fig.\ \ref{fig:objective-scatter} shows, for each of the (local) optima reached in these optimizations, the optimal learned noise variance $\sigma_n^2$ and the obtained values of the objective function at each local minimum.

For FITC, the learned noise variances $\sigma_n^2$ span five orders of magnitude, and many of them have very small values $\sim10^{-6}$, lying outside of the initialization interval, and clearly overfit on the data (see \cite[Figure 1]{bauer-understanding-sparse-GP-approximations}). Worse than that, the global optimum for FITC (red dot in left panel of Fig.\ \ref{fig:objective-scatter}) is found at the very small value of $\sigma_n^2\sim10^{-6}$, reproducing the findings of \cite[Section 3.1]{bauer-understanding-sparse-GP-approximations}. In contrast to that, our kl-PAC-SGP is much better behaved: the local optima have more reasonable $\sigma_n^2\in[2\cdot10^{-3},10^{-1}]$ and our global optimum has $\sigma_n^2\approx2.1\cdot10^{-2}$ (note however that the values of $\sigma_n$ learned by PAC-(S)GP will depend on the lengthscale $\varepsilon$ chosen for the loss function $\ell$; see also Table \ref{table-appendix-boston} in App.\ \ref{experimental-data-appendix}). While kl-PAC-SGP has further local optima at the small values $\sigma_n^2\in[10^{-5},2\cdot10^{-3}]$, where $\sigma_n^2$ does not move away from its small initialization value, these are easy to detect as the minimization objective attains the trivial value of $\approx1$.

This shows that our PAC-GP method is more stable than FITC on these pathological datasets and returns a more reasonable estimate of the noise level $\sigma_n^2$. It also reinforces the finding from the experiments in Sect.\ \ref{experiments-section} that our PAC-GP tends to underfit rather than overfit, hedging against violations of Theorem \ref{pac-bayes-theorem} and Eq.\ (\ref{pac-bayes-union-bound-eq}) by returning predictive GPs $Q$ of lower complexity $KL(Q\|P)$ by choosing larger $\sigma_n^2$.

\newpage

\section{Experiment: dependence of the upper bound on discretization}
\label{sec:exper-depend-upper}

\begin{figure}[h]
  \centering
\input{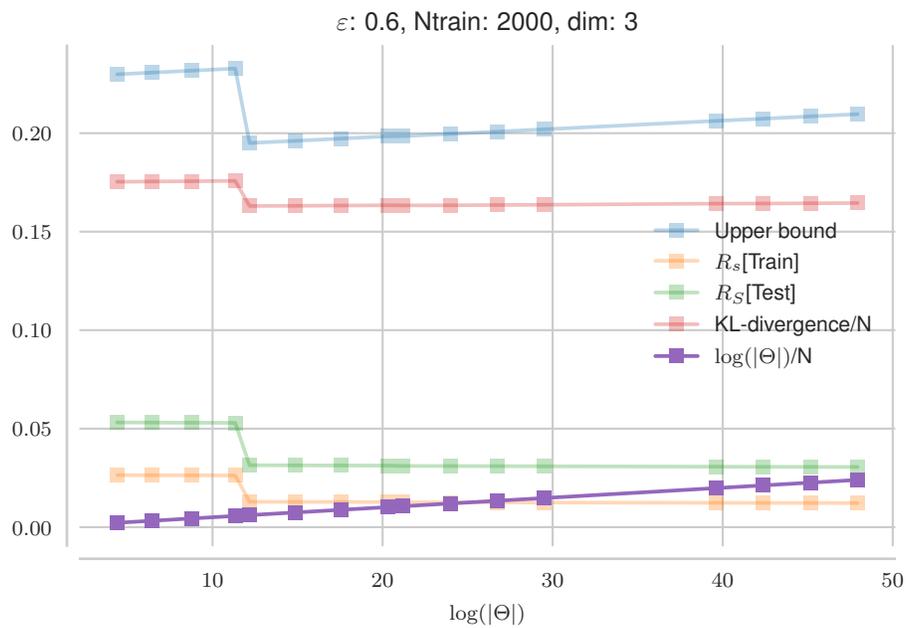}
  \caption{\textbf{Analysis of the discretization effect.} Upper bound (\ref{pac-bayes-union-bound-eq}) and its contributing terms, as well as the training and test risks, as a function of the discretization as measured by $\log|\Theta|$. Each line corresponds to the mean value over 68 iterations, when trained with our PAC-GP fitted to 3-dimensional data generated from an SE-ARD kernel with random lengthscales (see text, App. \ref{sec:exper-depend-upper}). }
  \label{fig:discretization-vs-log-theta}
\end{figure}

In order to assess the effect of discretizing hyperparameters $\theta$ (see Sections \ref{full-GP-section} and \ref{experiments-section}) on the performance of the resulting GP and on the upper bound, we ran our PAC-GP from Sect.\ \ref{full-GP-section} with different discretization settings and fitted them to artificial data. The results are shown in Fig.\ \ref{fig:discretization-vs-log-theta}.

Specifically, we generated inputs by uniformly sampling $x\in X:=[-3,3]^3\subset\mathbb R^3$. We sampled $N=2000$ training and $N=10000$ test outputs by sampling from a GP on the generated inputs, using an SE-ARD-kernel with randomly selected lengthscales for each of the $d=3$ dimensions. In more details, we sampled the kernel's log-lengthscales uniformly between $-1$ and $1$. To the generated data, we fitted a PAC-GP with a discretization given by $L\in \{1,2,4,8\}$ (see Sect.\ \ref{full-GP-section}) and a number of rounding digits $r\in\{0,1,2,4\}$ (i.e.\ $G=2L\cdot10^{r}$ in Sect.\ \ref{full-GP-section}). For example, for $L=1, r=0$, we only consider values $\log\theta\in\{-1,0,1\}$ resulting in $\log|\Theta|=(d+1)\cdot\ln(G+1)=4\ln3\approx4.4$.

To assess the contribution of the training risk $R_S$ and the $KL$-divergence term to the overall upper bound (\ref{pac-bayes-union-bound-eq}) on the generalization performance, we plotted the mean of each of these terms as a function of $\log|\Theta|$, averaged across 68 repetitions. Additionally, we plotted the risk on a test set to assess whether the actual test performance $R(Q)$ is affected by coarser discretization of the GP  hyperparameters. It can be seen in Fig.\ \ref{fig:discretization-vs-log-theta} that, as long as a minimal discrimination ability is allowed, both the training as well as the test risks are not affected by discretizing to a coarse grid of hyperparameters. Specifically, the jump that can be observed at $\ln|\Theta|\sim 11.3$ corresponds to going from $r=0$ to $r\geq1$, thereby keeping at least one decimal place in the discretization. We see that both the KL-divergence $KL(Q\|P)$ as well as the training risk $R_S(Q)$ is basically unaffected by the discretization for $r\geq1$, so any increase in the resulting upper bound is due to the increase in $\log|\Theta|$.

From this investigation, we find the discretization with $L=6$ and $r=2$ to be completely sufficient for accuracy, while the resulting $\ln|\Theta|$ term is still small compared to the contribution $KL(Q\|P)$ in the PAC-Bayes bound (\ref{pac-bayes-union-bound-eq}) as seen in our experiments (Sect.\ \ref{experiments-section}). For this discretization we have $\ln|\Theta|=(d+1)\ln(1201)\approx7.1(d+1)$ for an SE-ARD kernel in $d$ dimensions, and $\ln|\Theta|=2\ln(1201)\approx 14.2$ for a non-ARD SE-kernel. Note that -- for any fixed rounding accuracy of $\sim\log_2G$ bits -- the penalty term $\ln|\Theta|$ as well as the required storage capacity and computational effort all scale only linearly with the input dimension $d$; thus, our method requires the same computational complexity as other standard GP methods.

\newpage
\begin{landscape}
\section{Supplementary Tables}
\label{experimental-data-appendix}

\renewcommand{\thefootnote}{\fnsymbol{footnote}}

\begin{table}[h!]
	\caption{\textbf{Evaluation of full GP models (Fig.\ \ref{boston-fig}).} We compare our approach (``kl-PAC-GP'') with minimizing the looser bound $B_{Pin}$ (``sqrt-PAC GP'') and with the standard GP approach  (``full-GP'' \cite{rasmussen-williams-book}) on the following metrics (from left to right): upper bound $B$, Pinsker's upper bound $B_{Pin}$, Gibbs risk on the training data $R_S$[train], Gibbs risk on the test data $R_S$[test], mean squared error (MSE) on the test data, KL-divergence (normalized by the number of training samples, i.e.\ $KL(Q\|P)/N$), and the learned noise parameter $\sigma_n^2$. 
Shown are the averages$\pm$ standard errors over $10$ repetitions.}
	\label{table-appendix-boston}
	\centering
		\begin{tabular}{lclccccccc}
			\toprule
			\multicolumn{3}{c}{Model configuration} & \multicolumn{2}{c}{Upper bound} & \multicolumn{2}{c}{Gibbs risk} & \multicolumn{3}{c}{Model properties} \\
			\cmidrule(lr){1-3}
			\cmidrule(lr){4-5}
			\cmidrule(lr){6-7}
			\cmidrule(lr){8-10}
			dataset & epsilon & method & $B$ & $B_{Pin}$ & $R_S$[train] & $R_S$[test] & MSE& KL/N  & $\sigma^2$\\
			\midrule

  \multirow{15}{*}{boston} & 0.2 &kl-PAC-GP& 0.773 +/- 0.003& 0.798 +/- 0.003& 0.497 +/- 0.005& 0.536 +/- 0.006& 0.159 +/- 0.018& 0.126 +/- 0.003& 0.304 +/- 0.015\\
 & & sqrt-PAC-GP & 0.803 +/- 0.016& 0.834 +/- 0.019& 0.573 +/- 0.039& 0.599 +/- 0.034& 0.420 +/- 0.129& 0.087 +/- 0.018& 1334.991 +/- 1334.032\footnotemark[9]\\
 & & full-GP & 0.809 +/- 0.004& 0.851 +/- 0.006& 0.372 +/- 0.012& 0.501 +/- 0.007& 0.114 +/- 0.015& 0.405 +/- 0.023& 0.066 +/- 0.004\\ 
  \cmidrule(lr){2-10}
  &0.4 &kl-PAC-GP& 0.498 +/- 0.004& 0.507 +/- 0.003& 0.211 +/- 0.005& 0.243 +/- 0.008& 0.161 +/- 0.018& 0.120 +/- 0.002& 0.328 +/- 0.013\\
  && sqrt-PAC-GP & 0.498 +/- 0.004& 0.507 +/- 0.003& 0.218 +/- 0.004& 0.250 +/- 0.008& 0.165 +/- 0.019& 0.111 +/- 0.002& 0.371 +/- 0.013\\
  && full-GP & 0.548 +/- 0.005& 0.576 +/- 0.006& 0.097 +/- 0.008& 0.217 +/- 0.007& 0.114 +/- 0.015& 0.405 +/- 0.023& 0.066 +/- 0.004\\
  \cmidrule(lr){2-10}
   &0.6 &kl-PAC-GP& 0.333 +/- 0.004& 0.376 +/- 0.002& 0.093 +/- 0.003& 0.115 +/- 0.006& 0.167 +/- 0.019& 0.104 +/- 0.002& 0.424 +/- 0.015\\
   && sqrt-PAC-GP & 0.336 +/- 0.003& 0.373 +/- 0.002& 0.111 +/- 0.003& 0.133 +/- 0.006& 0.182 +/- 0.020& 0.082 +/- 0.001& 0.625 +/- 0.014\\
   && full-GP & 0.432 +/- 0.009& 0.503 +/- 0.010& 0.025 +/- 0.003& 0.096 +/- 0.008& 0.114 +/- 0.015& 0.405 +/- 0.023& 0.066 +/- 0.004\\
   \cmidrule(lr){2-10}
   &0.8 &kl-PAC-GP& 0.247 +/- 0.003& 0.313 +/- 0.002& 0.053 +/- 0.002& 0.069 +/- 0.005& 0.181 +/- 0.019& 0.080 +/- 0.002& 0.666 +/- 0.017\\
   && sqrt-PAC-GP & 0.253 +/- 0.003& 0.308 +/- 0.002& 0.072 +/- 0.002& 0.089 +/- 0.006& 0.212 +/- 0.022& 0.055 +/- 0.001& 1.163 +/- 0.020\\
   && full-GP & 0.394 +/- 0.011& 0.486 +/- 0.011& 0.008 +/- 0.001& 0.046 +/- 0.006& 0.114 +/- 0.015& 0.405 +/- 0.023& 0.066 +/- 0.004\\
   \cmidrule(lr){2-10}
   &1.0 &kl-PAC-GP& 0.198 +/- 0.002& 0.278 +/- 0.001& 0.035 +/- 0.002& 0.047 +/- 0.004& 0.196 +/- 0.020& 0.062 +/- 0.001& 1.025 +/- 0.028\\
   && sqrt-PAC-GP & 0.206 +/- 0.002& 0.271 +/- 0.001& 0.052 +/- 0.001& 0.066 +/- 0.005& 0.238 +/- 0.023& 0.040 +/- 0.001& 1.987 +/- 0.036\\
   && full-GP & 0.379 +/- 0.013& 0.481 +/- 0.012& 0.003 +/- 0.000& 0.026 +/- 0.004& 0.114 +/- 0.015& 0.405 +/- 0.023& 0.066 +/- 0.004 \\
			\bottomrule
		\end{tabular}
	\vskip -0.15in
\end{table}

\footnotetext[9]{One of the 10 iterations ended up in a local optimum with very large $\sigma_n^2$. In contrast to overfitting, this corresponds to underfitting as can be seen by the small value of $KL/N$. Observe also that, within the setting $\varepsilon=0.2$, the upper bound $B$ is close to $1$ for all GPs, indicating a hard prediction problem for the given accuracy of $\varepsilon=0.2$.}

\bigskip
\bigskip

\begin{table}[h!]
	\caption{\textbf{Evaluation of sparse GP models (Fig.\ \ref{sparse-fig})}.
	We benchmark our method  (``kl-PAC-SGP'') against minimizing the looser bound $B_{Pin}$ (``sqrt-PAC-SGP'') and two standard sparse GP approaches (VFE \cite{titsias09} and VFE \cite{snelson-spgp}) using the following criteria (from left to right): upper bound $B$, Pinsker's upper bound $B_{Pin}$, Gibbs risk on the training data $R_S$[train], Gibbs risk on the test data $R_S$[test], mean squared error (MSE) on the test data, KL-divergence (normalized by the number of training samples, i.e.\ $KL(Q\|P)/N$), and the learned noise parameter $\sigma_n^2$.
	The number of inducing inputs is fixed to $M=500$, and we use the 0-1-loss function $\ell(y,\widehat y)=\ii_{\widehat y\notin[y-\varepsilon,y+\varepsilon]}$ with $\varepsilon=0.6$ (see Sect.\ \ref{experiments-section}).
	Automatic feature determination (ARD) is beneficial on the datasets \textit{pol} and \textit{kin40k} and has no effect on \textit{sarcos}.
	We report mean values $\pm$ standard errors over $10$ iterations.}
	\label{table-appendix-sparse-GPs}
	\centering

		\begin{tabular}{llcccccccc}
			\toprule
			\multicolumn{3}{c}{Model configuration} & \multicolumn{2}{c}{Upper bound} & \multicolumn{2}{c}{Gibbs risk} & \multicolumn{3}{c}{Model properties} \\
			\cmidrule(lr){1-3}
			\cmidrule(lr){4-5}
			\cmidrule(lr){6-7}
			\cmidrule(lr){8-10}
			dataset & method & ARD & $B$ & $B_{Pin}$ & $R_S$[train] & $R_S$[test] & MSE& KL/N & $\sigma^2$ \\
			\midrule

\multirow{8}{*}{pol} &kl-PAC-SGP& \xmark & 0.217 +/- 0.001& 0.252 +/- 0.000& 0.106 +/- 0.001& 0.115 +/- 0.001& 0.114 +/- 0.001& 0.041 +/- 0.000& 0.316 +/- 0.015\\
& sqrt-PAC-SGP & \xmark& 0.221 +/- 0.001& 0.248 +/- 0.000& 0.126 +/- 0.001& 0.133 +/- 0.001& 0.124 +/- 0.001& 0.028 +/- 0.000& 0.626 +/- 0.025\\
& VFE & \xmark& 0.257 +/- 0.000& 0.312 +/- 0.000& 0.071 +/- 0.000& 0.081 +/- 0.001& 0.090 +/- 0.001& 0.114 +/- 0.000& 0.102 +/- 0.000\\
& FITC & \xmark& 0.359 +/- 0.001& 0.384 +/- 0.001& 0.149 +/- 0.001& 0.160 +/- 0.001& 0.092 +/- 0.001& 0.109 +/- 0.002& 0.000 +/- 0.000 \\
			\cmidrule(lr){2-10}
&kl-PAC-SGP& \checkmark& 0.083 +/- 0.000& 0.172 +/- 0.000& 0.011 +/- 0.000& 0.015 +/- 0.000& 0.036 +/- 0.000& 0.035 +/- 0.000& 0.187 +/- 0.003\\
& sqrt-PAC-SGP & \checkmark& 0.094 +/- 0.000& 0.159 +/- 0.000& 0.029 +/- 0.000& 0.032 +/- 0.000& 0.044 +/- 0.000& 0.017 +/- 0.000& 0.825 +/- 0.011\\
& VFE & \checkmark& 0.198 +/- 0.000& 0.324 +/- 0.000& 0.002 +/- 0.000& 0.006 +/- 0.000& 0.015 +/- 0.000& 0.190 +/- 0.000& 0.016 +/- 0.000\\
& FITC & \checkmark& 0.247 +/- 0.001& 0.333 +/- 0.001& 0.029 +/- 0.000& 0.032 +/- 0.001& 0.027 +/- 0.000& 0.168 +/- 0.001& 0.000 +/- 0.000 \\
			\midrule
\multirow{8}{*}{sarcos} &kl-PAC-SGP&\xmark & 0.031 +/- 0.000& 0.083 +/- 0.000& 0.009 +/- 0.000& 0.010 +/- 0.000& 0.033 +/- 0.000& 0.010 +/- 0.000& 0.526 +/- 0.004\\
& sqrt-PAC-SGP & \xmark& 0.038 +/- 0.000& 0.066 +/- 0.000& 0.023 +/- 0.000& 0.023 +/- 0.000& 0.044 +/- 0.000& 0.003 +/- 0.000& 3.600 +/- 0.006\\
& VFE &\xmark & 0.097 +/- 0.000& 0.215 +/- 0.000& 0.002 +/- 0.000& 0.003 +/- 0.000& 0.017 +/- 0.000& 0.090 +/- 0.000& 0.019 +/- 0.000\\
& FITC &\xmark & 0.116 +/- 0.000& 0.211 +/- 0.000& 0.014 +/- 0.000& 0.015 +/- 0.000& 0.019 +/- 0.000& 0.076 +/- 0.000& 0.000 +/- 0.000\\
			\cmidrule(lr){2-10}

 &kl-PAC-SGP& \checkmark& 0.031 +/- 0.000& 0.095 +/- 0.000& 0.005 +/- 0.000& 0.007 +/- 0.000& 0.029 +/- 0.000& 0.012 +/- 0.000& 0.389 +/- 0.002\\
& sqrt-PAC-SGP & \checkmark& 0.039 +/- 0.000& 0.079 +/- 0.000& 0.018 +/- 0.000& 0.018 +/- 0.000& 0.040 +/- 0.000& 0.003 +/- 0.000& 2.682 +/- 0.009\\
& VFE & \checkmark& 0.092 +/- 0.000& 0.212 +/- 0.000& 0.002 +/- 0.000& 0.002 +/- 0.000& 0.016 +/- 0.000& 0.084 +/- 0.000& 0.017 +/- 0.000\\
& FITC & \checkmark& 0.115 +/- 0.000& 0.215 +/- 0.000& 0.012 +/- 0.000& 0.012 +/- 0.000& 0.017 +/- 0.000& 0.079 +/- 0.000& 0.000 +/- 0.000 \\
			\midrule

\multirow{8}{*}{kin40k} & kl-PAC-SGP& \xmark & 0.154 +/- 0.000& 0.219 +/- 0.000& 0.045 +/- 0.000& 0.053 +/- 0.000& 0.059 +/- 0.001& 0.059 +/- 0.000& 0.262 +/- 0.014\\
& sqrt-PAC-SGP & \xmark & 0.162 +/- 0.001& 0.207 +/- 0.001& 0.071 +/- 0.000& 0.079 +/- 0.001& 0.082 +/- 0.001& 0.036 +/- 0.000& 0.658 +/- 0.046\\
 &VFE &\xmark & 0.238 +/- 0.000& 0.341 +/- 0.000& 0.014 +/- 0.000& 0.019 +/- 0.000& 0.030 +/- 0.000& 0.212 +/- 0.000& 0.040 +/- 0.000\\
& FITC & \xmark & 0.302 +/- 0.001& 0.359 +/- 0.001& 0.066 +/- 0.001& 0.068 +/- 0.001& 0.082 +/- 0.003& 0.171 +/- 0.002& 0.000 +/- 0.000 \\
			\cmidrule(lr){2-10}

&kl-PAC-SGP& \checkmark& 0.115 +/- 0.000& 0.190 +/- 0.000& 0.028 +/- 0.000& 0.034 +/- 0.000& 0.049 +/- 0.000& 0.050 +/- 0.000& 0.254 +/- 0.012\\
& sqrt-PAC-SGP & \checkmark& 0.126 +/- 0.000& 0.175 +/- 0.000& 0.054 +/- 0.000& 0.059 +/- 0.001& 0.071 +/- 0.000& 0.027 +/- 0.000& 0.814 +/- 0.013\\
& VFE & \checkmark& 0.212 +/- 0.000& 0.327 +/- 0.000& 0.007 +/- 0.000& 0.011 +/- 0.000& 0.024 +/- 0.000& 0.202 +/- 0.000& 0.031 +/- 0.000\\
& FITC & \checkmark& 0.277 +/- 0.000& 0.347 +/- 0.000& 0.046 +/- 0.000& 0.048 +/- 0.000& 0.053 +/- 0.001& 0.179 +/- 0.000& 0.000 +/- 0.000\\
			\bottomrule
		\end{tabular}

	\vskip -0.15in
\end{table}

\begin{table}[h!]
	\caption{\textbf{Evaluation of inverted Gaussian as loss function $\ell_{\exp}$}.
Using the more distance-sensitive loss function $\ell_{\exp}$ from Eq.\ (\ref{loss-function-exp}) for our methods ``kl-PAC-SGP'' and ``sqrt-PAC-SGP'', we run them against the two standard sparse GP approaches (VFE \cite{titsias09} and VFE \cite{snelson-spgp}) for our three sparse-GP datasets, see Sect.\ \ref{experiments-section}. Compare also to Table \ref{table-appendix-sparse-GPs}, where the same investigation was done using the 0-1-loss $\ell$ (here, we only report the favorable ARD/non-ARD settings displayed in Fig.\ \ref{sparse-fig}, cf.\ Table \ref{table-appendix-sparse-GPs}). We again use the following criteria (from left to right): upper bound $B$, Pinsker's upper bound $B_{Pin}$, Gibbs risk on the training data $R_S$[train], Gibbs risk on the test data $R_S$[test], mean squared error (MSE) on the test data, KL-divergence (normalized by the number of training samples, i.e.\ $KL(Q\|P)/N$), and the learned noise parameter $\sigma_n^2$.
	The number of inducing inputs is fixed to $M=500$. We report mean values $\pm$ standard errors over $10$ iterations.}
	\label{table-appendix-inverted-Gaussian-loss}
	\centering
		\begin{tabular}{llcccccccc}
			\toprule
			\multicolumn{3}{c}{Model configuration} & \multicolumn{2}{c}{Upper bound} & \multicolumn{2}{c}{Gibbs risk} & \multicolumn{3}{c}{Model properties} \\
			\cmidrule(lr){1-3}
			\cmidrule(lr){4-5}
			\cmidrule(lr){6-7}
			\cmidrule(lr){8-10}
			dataset & method & ARD & $B$ & $B_{Pin}$ & $R_S$[train] & $R_S$[test] & MSE& KL/N & $\sigma^2$ \\
			\midrule
			\multirow{4}{*}{pol} &kl-PAC-SGP& \checkmark& 0.199 +/- 0.000& 0.247 +/- 0.000& 0.077 +/- 0.000& 0.019 +/- 0.000& 0.027 +/- 0.000& 0.041 +/- 0.000& 0.216 +/- 0.003\\& sqrt-PAC-SGP & \checkmark& 0.208 +/- 0.001& 0.245 +/- 0.000& 0.100 +/- 0.001& 0.031 +/- 0.001& 0.040 +/- 0.001& 0.025 +/- 0.000& 0.461 +/- 0.006\\
& VFE & \checkmark& 0.288 +/- 0.000& 0.361 +/- 0.000& 0.041 +/- 0.000& 0.006 +/- 0.000& 0.015 +/- 0.000& 0.189 +/- 0.000& 0.016 +/- 0.000\\
& FITC & \checkmark& 0.345 +/- 0.001& 0.390 +/- 0.001& 0.085 +/- 0.000& 0.030 +/- 0.001& 0.027 +/- 0.000& 0.169 +/- 0.001& 0.000 +/- 0.000 \\
			\midrule
			\multirow{4}{*}{sarcos} &kl-PAC-SGP& \xmark & 0.116 +/- 0.000& 0.144 +/- 0.000& 0.073 +/- 0.000& 0.012 +/- 0.000& 0.032 +/- 0.000& 0.009 +/- 0.000& 0.213 +/- 0.001\\
& sqrt-PAC-SGP & \xmark & 0.119 +/- 0.000& 0.139 +/- 0.000& 0.086 +/- 0.000& 0.017 +/- 0.000& 0.039 +/- 0.000& 0.005 +/- 0.000& 0.611 +/- 0.002\\
& VFE & \xmark & 0.190 +/- 0.000& 0.259 +/- 0.000& 0.047 +/- 0.000& 0.003 +/- 0.000& 0.017 +/- 0.000& 0.090 +/- 0.000& 0.019 +/- 0.000\\
& FITC &\xmark & 0.227 +/- 0.000& 0.276 +/- 0.000& 0.079 +/- 0.000& 0.015 +/- 0.000& 0.019 +/- 0.000& 0.077 +/- 0.000& 0.000 +/- 0.000 \\
			\midrule
			\multirow{4}{*}{kin40k} &kl-PAC-SGP& \checkmark& 0.260 +/- 0.001& 0.290 +/- 0.000& 0.127 +/- 0.001& 0.039 +/- 0.001& 0.055 +/- 0.000& 0.051 +/- 0.000& 0.124 +/- 0.014\\& sqrt-PAC-SGP & \checkmark& 0.262 +/- 0.001& 0.287 +/- 0.001& 0.142 +/- 0.001& 0.048 +/- 0.001& 0.061 +/- 0.000& 0.040 +/- 0.000& 0.232 +/- 0.017\\
& VFE & \checkmark& 0.347 +/- 0.000& 0.396 +/- 0.000& 0.076 +/- 0.000& 0.011 +/- 0.000& 0.024 +/- 0.000& 0.202 +/- 0.000& 0.031 +/- 0.000\\
& FITC & \checkmark& 0.379 +/- 0.000& 0.413 +/- 0.000& 0.112 +/- 0.000& 0.048 +/- 0.000& 0.053 +/- 0.001& 0.179 +/- 0.001& 0.000 +/- 0.000 \\
			\bottomrule
		\end{tabular}
	\vskip -0.15in
\end{table}

\end{landscape}




\begin{thebibliography}{99}

\bibitem{rasmussen-williams-book}
C.\ E.\ Rasmussen, C.\ K.\ I.\ Williams, ``Gaussian Processes for Machine Learning'', The MIT Press (2006).

\bibitem{bauer-understanding-sparse-GP-approximations}
M.\ Bauer, M.\ v.\ d.\ Wilk, C.\ E.\ Rasmussen, ``Understanding Probabilistic Sparse Gaussian Process Approximations'', In NIPS (2016).

\bibitem{understanding-machine-learning-book}
S.\ Shalev-Shwartz, S.\ Ben-David, ``Understanding Machine Learning: From Theory to Algorithms'', Cambridge University Press (2014).

\bibitem{understanding-deep-learning-requires-rethinking-generalization}
C.\ Zhang, S.\ Bengio, M.\ Hardt, B.\ Recht, O.\ Vinyals, ``Understanding deep learning requires rethinking generalization'', In ICLR (2017).

\bibitem{jordan-ghahramani-et-al-vi}
M.\ Jordan, Z.\ Ghahramani, T.\ Jaakkola, L.\ Saul, ``Introduction to variational methods for graphical models'', Machine Learning, 37, 183-233 (1999).

\bibitem{titsias09}
M.\ Titsias, ``Variational Learning of Inducing Variables in Sparse Gaussian Processes'', In AISTATS (2009).

\bibitem{mcallester-pac-bayesian-model-averaging}
D.\ McAllester, ``PAC-Bayesian model averaging'', COLT (1999).

\bibitem{asdf2}
D.\ McAllester, ``PAC-Bayesian Stochastic Model Selection'', Machine Learning 51, 5-21 (2003).

\bibitem{catoni-thermodyn-of-ml}
O.\ Catoni, ``Pac-Bayesian Supervised Classification: The Thermodynamics of Statistical Learning'', IMS Lecture Notes Monograph Series 2007, Vol. 56 (2007).

\bibitem{seeger-classification-paper}
M.\ Seeger, ``PAC-Bayesian Generalization Error Bounds for Gaussian Process Classification'', Journal of Machine Learning Research 3, 233-269 (2002).


\bibitem{tighter-pac-bayes-bounds-svm}
A.\ Ambroladze, E.\ Parrado-Hernández, J.\ Shawe-Taylor, ``Tighter PAC-Bayes bounds'', In NIPS (2007).

\bibitem{pac-bayes-and-margins}
J.\ Langford, J.\ Shawe-Tayor, ``PAC-Bayes \& margins'', In NIPS (2002).

\bibitem{pac-bayes-learning-of-linear-classifiers}
P.\ Germain, A.\ Lacasse, F.\ Laviolette, M.\ Marchand, ``PAC-Bayesian Learning of Linear Classifiers'', In ICML (2009).

\bibitem{nonvacuous-generalization-bounds-snn}
G.\ K.\ Dziugaite, D.\ M.\ Roy, ``Computing Nonvacuous Generalization Bounds for Deep (Stochastic) Neural Networks with Many More Parameters than Training Data'', In UAI (2017).

\bibitem{snelson-spgp}
E.\ Snelson, Z.\ Ghahramani, ``Sparse Gaussian Processes using Pseudo-inputs'', In NIPS (2005).

\bibitem{seeger-dtc}
M.\ Seeger, C.\ K.\ I.\ Williams, N.\ Lawrence, ``Fast Forward Selection to Seepd Up Sparse Gaussian Process Regression'', In AISTATS (2003).

\bibitem{pac-bayesian-theory-meets-bayesian-inference}
P.\ Germain, F.\ Bach, A.\ Lacoste, S.\ Lacoste-Julien, ``PAC-Bayesian Theory Meets Bayesian Inference'', In NIPS (2016).

\bibitem{excess-risk-bounds-using-vi-in-gaussian-models}
R.\ Sheth, R.\ Khardon, ``Excess Risk Bounds for the Bayes Risk using Variational Inference in Latent Gaussian Models'', In NIPS (2017).

\bibitem{vapnik-the-nature-of-statistical-learning-theory}
V.\ Vapnik, ``The Nature of Statistical Learning Theory'', Springer (1995).

\bibitem{maurer-note-pac-bayes}
A.\ Maurer, ``A Note on the PAC Bayesian Theorem'', arXiv:cs/0411099 (2004).

\bibitem{pac-bayesian-bounds-based-on-the-renyi-divergence}
L.\ Begin, P.\ Germain, F.\ Laviolette, J.-F.\ Roy, ``PAC-Bayesian Bounds based on the Renyi Divergence'', In AISTATS (2016).

\bibitem{kullback-leibler-paper}
S.\ Kullback, R.\ Leibler, ``On information and sufficiency'', Annals of Mathematical Statistics 22, 79-86 (1951).

\bibitem{KL-divergence-between-stochastic-processes}
A.\ G.\ de G.\ Matthews, J.\ Hensman, R.\ Turner, Z.\ Ghahramani, ``On Sparse Variational Methods and the Kullback-Leibler Divergence between Stochastic Processes'', In AISTATS (2016).

\bibitem{unifying-view-sparse-gp-approx}
J.\ Quinonero-Candela, C.\ E.\ Rasmussen, ``A Unifying View of Sparse Approximate Gaussian Process Regression'', Journal of Machine Learning Research 6, 1939-1959 (2005).

\bibitem{bui-yan-turner-unifying}
T.\ D.\ Bui, J.\ Yan, R.\ E.\ Turner, ``A Unifying Framework for Gaussian Process Pseudo-Point Approximations using Power Expectation Propagation'', Journal of Machine Learning Research 18, 1-72 (2017).

\bibitem{scalable-variational-gaussian-process-classification}
J.\ Hensman, A.\ Matthews, Z.\ Ghahramani, ``Scalable Variational Gaussian Process Classification'', In AISTATS (2015).

\bibitem{gpflow}
A.\ Matthews, M.\ van der Wilk, T.\ Nickson, K.\ Fujii, A.\ Boukouvalas, P.\ {Le{\'o}n-Villagr{\'a}}, Z.\ Ghahramani, J.\ Hensman, ``{{GP}flow: A {G}aussian process library using {T}ensor{F}low}'', Journal of Machine Learning Research 18, 1-6 (2017).

\bibitem{tensorflow}
M.\ Abadi {\it{et al.}}, ``{TensorFlow}: Large-Scale Machine Learning on Heterogeneous Systems'', \url{https://www.tensorflow.org/} (2015).

\bibitem{more-general-than-free-form}
C.-A.\ Cheng, B.\ Boots, ``Variational Inference for Gaussian Process Models with Linear Complexity'', In NIPS (2017).









\end{thebibliography}
\end{document}